\documentclass[10pt,twocolumn,letterpaper]{article}

\usepackage[pagenumbers]{cvpr} 

\newif\ifdrafting
\draftingtrue 
\ifdrafting
    \newcommand{\cy}[1]{\textcolor{blue}{CY: #1}}
    \newcommand{\yx}[1]{\textcolor{orange}{YX: #1}}
    \newcommand{\VJ}[1]{{\color{magenta}[VJ: #1]}}
    
    \newcommand{\todo}[1]{{\color{red}{#1}}}
    \newcommand{\TODO}[1]{\textbf{\color{red}[TODO: #1]}}
    \newcommand{\newadd}[1]{\textcolor{black}{#1}}
\else
    \newcommand{\cy} [1] {}
    \newcommand{\yx} [1] {}
    \newcommand{\VJ} [1] {}
    \newcommand{\todo} [1] {}
    \newcommand{\newadd} [1] {}
    \newcommand{\TODO}[1]{}
\fi

\def\ie{\emph{i.e}}

\newcommand{\inlinesection}[1]{\vspace{1mm} \noindent {\bf #1}}

\newcommand{\urlNewWindow}[1]{\href[pdfnewwindow=true]{#1}{\nolinkurl{#1}}}

\newcommand{\nameold}{SV4D\xspace}
\newcommand{\name}{SV4D 2.0\xspace}

\newcommand{\Times}{{\mkern-2mu\times\mkern-2mu}}

\definecolor{rblue}{RGB}{68, 114, 196}
\definecolor{rgreen}{RGB}{112, 173, 71}
\definecolor{rorange}{RGB}{237, 125, 49}
\definecolor{rred}{RGB}{255, 0, 0}

\definecolor{cvprblue}{rgb}{0.21,0.49,0.74}
\usepackage[pagebackref,breaklinks,colorlinks,allcolors=cvprblue]{hyperref}

\usepackage[utf8]{inputenc} 
\usepackage[T1]{fontenc}    
\usepackage{url}            
\usepackage{booktabs}       
\usepackage{amsfonts}       
\usepackage{nicefrac}       
\usepackage{microtype}      
\usepackage{xcolor}         
\usepackage{bm,mathtools}
\usepackage{multirow}
\usepackage{floatrow}
\usepackage{float}
\floatstyle{plaintop}
\restylefloat{table}
\usepackage{caption}
\usepackage{enumitem}
\usepackage{booktabs}
\usepackage{graphicx}
\usepackage{natbib}
\usepackage{soul}
\usepackage{colortbl}
\usepackage{gensymb}

\def\paperID{26} 
\def\confName{CVPR}
\def\confYear{2025}

\title{SV4D 2.0: Enhancing Spatio-Temporal Consistency in Multi-View Video Diffusion for High-Quality 4D Generation}

\author{%
  Chun-Han Yao$^{1*}$ \quad
  Yiming Xie$^{1,2*}$ \quad
  Vikram Voleti$^1$ \quad
  Huaizu Jiang$^{2\dagger}$ \quad
  Varun Jampani$^{1\dagger}$
  \\
  $^1$ Stability AI \quad
  $^2$ Northeastern University\\
  $^*$ Equal contribution \quad
  $^\dagger$ Equal advising
  \\
}

\begin{document}

\twocolumn[{%
\renewcommand\twocolumn[1][]{#1}%
\maketitle
\includegraphics[width=.95\linewidth]{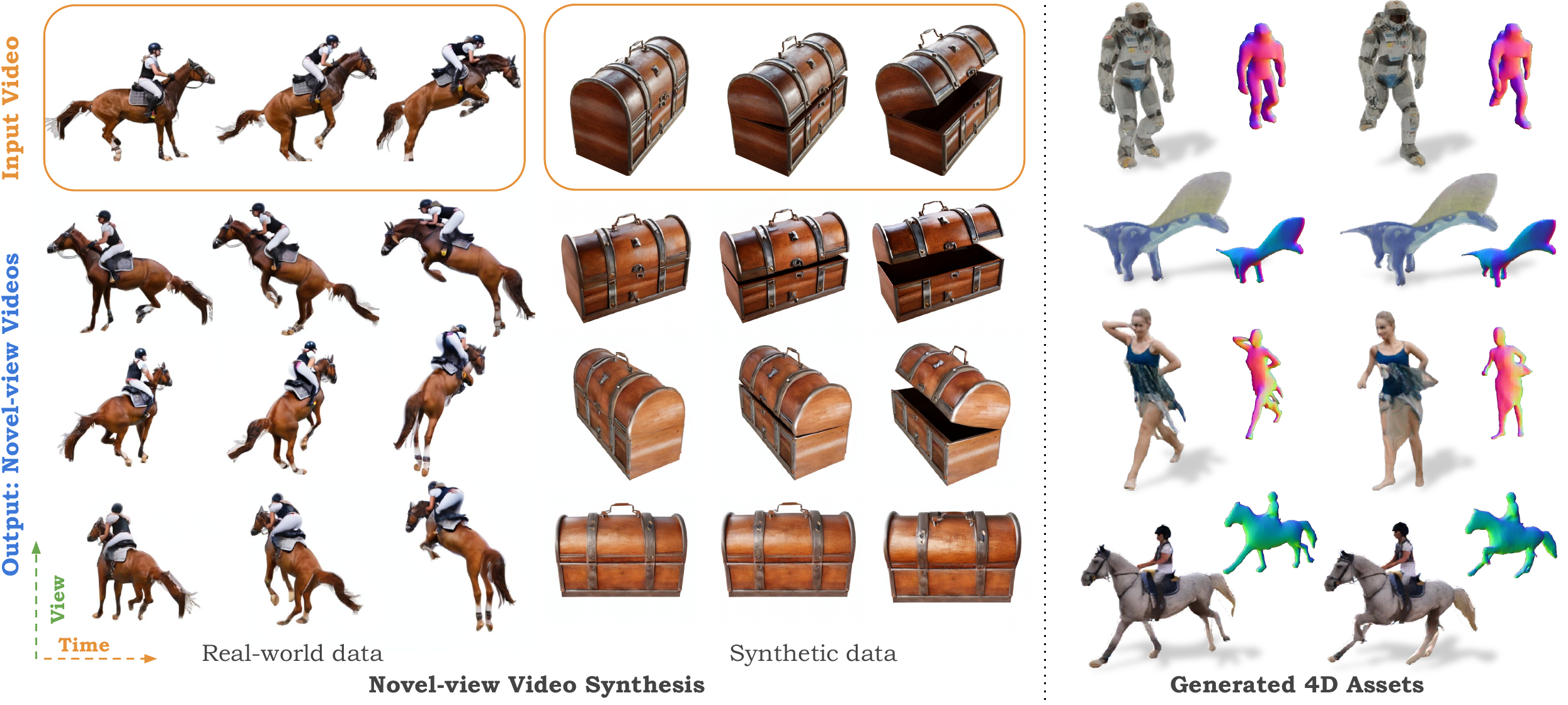}
    \vspace{-2mm}
    \captionof{figure}{
        {\bf \name} generates 
        multiple novel-view videos from an input monocular video. 
        The generated novel-view videos have high fidelity in terms of detail sharpness and consistency across view and time/frame axes, which can be used to optimize high-quality 4D assets. 
    }
    \vspace{15pt}
\label{fig:teaser}
}]

\begin{abstract}
We present Stable Video 4D 2.0 (SV4D 2.0), a multi-view video diffusion model for dynamic 3D asset generation. Compared to its predecessor SV4D~\cite{xie2024sv4d}, SV4D 2.0 is more robust to occlusions and large motion, generalizes better to real-world videos, and produces higher-quality outputs in terms of detail sharpness and spatio-temporal consistency. We achieve this by introducing key improvements in multiple aspects: 1) network architecture: eliminating the dependency of reference multi-views and designing blending mechanism for 3D and frame attention, 2) data: enhancing quality and quantity of training data, 3) training strategy: adopting progressive 3D-4D training for better generalization, and 4) 4D optimization: handling 3D inconsistency and large motion via 2-stage refinement and progressive frame sampling. Extensive experiments demonstrate significant performance gain by SV4D 2.0 both visually and quantitatively, achieving better detail (-14\% LPIPS) and 4D consistency (-44\% FV4D) in novel-view video synthesis and 4D optimization (-12\% LPIPS and -24\% FV4D) compared to SV4D. Project page: \urlNewWindow{https://sv4d20.github.io}.
\end{abstract}

\vspace{-4mm}
\section{Introduction}
\vspace{-2mm}

Dynamic 3D generation, also known as 4D generation, aims to synthesize a moving object/scene in 3D space, capturing the dynamics of the subject across time.
The generated 4D assets can be used in a wide range of applications such as video games, movie production, AR/VR experience, etc.
Traditionally, such 4D asset generation requires a tedious process where 3D artists handcraft 3D geometry, texture, as well as rigging on the 3D object. 
With the recent advances in image~\cite{rombach2022high,ruiz2023dreambooth} and video~\cite{blattmann2023align,blattmann2023stable,voleti2022mcvd,girdhar2023emu} diffusion models which show astonishing generation quality in the 2D domain, there has been a growing interest in extending them to 3D and 4D to speed up and/or automate this process.

Generating a dynamic 3D object from a single-view video is a highly ill-posed problem due to the ambiguity of object appearance and motion at unseen camera views.
Moreover, considering the lack of large-scale 4D datasets, and the requirement of high model capacity 
to handle the high-dimensional nature of 4D, it is challenging to train a native 4D generative model that can generalize to diverse object categories and motion.
A common way to tackle this problem using diffusion models is to directly optimize a 4D asset using the priors in pre-trained video and multi-view generative models via a score-distillation sampling (SDS) loss~\citep{singer2023text4d,bah20244dfy,ling2023alignyourgaussians,ren2023dreamgaussian4d,zhao2023animate124,jiang2023consistent4d,zeng2024stag4d,yin20234dgen}.
While proven to be effective in some cases, SDS loss can easily lead to artifacts such as spatio-temporal inconsistency or over-saturated color if not handled properly.
Several recent works~\citep{yang2024diffusion,sun2024eg4d,xie2024sv4d} have shown that reconstruction from synthesized videos is sufficient to produce high-quality 4D assets without SDS loss, by sampling videos of the subject at novel viewpoints via novel-view video synthesis (NVVS) techniques and using them as photogrammetry pseudo-ground truths.

The main challenge for NVVS is to generate multi-view videos that are consistent spatially (\ie, across novel views), and temporally (\ie, in dynamic motion). 
Earlier works~\cite{yang2024diffusion,sun2024eg4d} on NVVS use separate models for novel-view synthesis and dynamic object motion, which often causes blurry details or poor spatio-temporal consistency, and thus can only handle small object motion.
Some recent methods~\cite{xie2024sv4d,li2024vividzoo,zhang20244diffusion,liang2024diffusion4d} jointly learn spatial and temporal consistency in a single NVS model by repurposing and finetuning a video diffusion model on synthetic 4D data~\cite{deitke2023objaverse,deitke2023objaversexl}.
However, their success is limited to low resolution~\cite{li2024vividzoo,zhang20244diffusion,liang2024diffusion4d}, small object motion~\cite{liang2024diffusion4d}, or short video length~\cite{xie2024sv4d}.

In this work, we propose Stable Video 4D 2.0 (SV4D 2.0), a NVVS diffusion model that generates high-resolution multi-view videos of a dynamic 3D object, given a monocular video and a user-specified camera trajectory as input. 
We then use the NVVS outputs of SV4D 2.0 to optimize a 4D asset.
In SV4D 2.0, we build upon the SV4D~\cite{xie2024sv4d} framework and make several key modifications to improve 1) the ability to generate longer videos and sparser (\ie, more distant) views 
, 2) robustness to self-occlusion and large motion, and 3) output quality in terms of details and spatio-temporal consistency.
Our improvements over SV4D include: 
\begin{itemize}
\item \textbf{\textit{Network architecture}}:
We make several modifications to the architecture, including 3D attention layers and disentangled $\alpha$-blending of 3D and temporal information, enabling sparse novel-view synthesis and thus longer video generation in a single inference pass. Unlike SV4D which requires SV3D~\cite{voleti2024sv3d} to generate reference multi-views, we remove this dependency using a random masking strategy, resulting in robustness to self-occlusion in the first frame. 
This also enables more flexible inference sampling.
\item \textbf{\textit{Data Curation}}: 
We improve the quality of our synthetic training dataset by disentangling global transformation from local motion, rectifying off-center objects, and filtering out objects with small motion or inconsistent scaling. 
\item \textbf{\textit{Training strategy}}:
Unlike most 4D generation methods~\cite{yang2024diffusion,liang2024diffusion4d,ren2024l4gm,li2024vividzoo} which directly finetune a modified video or 3D generation model on 4D data, we adopt a progressive 3D-to-4D training.
This smoothens the adaption to new network architecture and bridges the 3D-4D domain gap
, which helps preserves the rich video and 3D priors and thus generalizes better to in-the-wild videos.
\item \textbf{\textit{4D optimization}}: 
We propose a novel 2-stage optimization scheme for a 4D asset with visibility-weighted losses and progressive frame sampling to handle view inconsistencies in the synthesized novel-view videos and large object motion, respectively. This leads to improved 4D reconstruction quality compared to prior works.
\end{itemize}

We perform comprehensive evaluations of NVVS and 4D optimization 
on both synthetic datasets (ObjaverseDy~\cite{xie2024sv4d} and Consistent4D~\cite{jiang2023consistent4d}) and real-world videos (DAVIS~\cite{Perazzi_CVPR_2016,Pont-Tuset_arXiv_2017,Caelles_arXiv_2019}).
As shown in~\cref{fig:teaser}, SV4D 2.0 outputs demonstrate high image quality, spatio-temporal consistency, and fidelity to the input videos. 
Despite being only trained on the synthetic 3D and 4D object datasets, SV4D 2.0 also demonstrates good generalization on real-world videos.
Quantitatively, SV4D 2.0 achieves better details (-14\% LPIPS) and 4D consistency (-44\% FV4D) in video synthesis and -12\% LPIPS and -24\% FV4D in 4D optimization compared to SV4D, setting a new state-of-the-art for 4D generation.

\vspace{-2mm}
\section{Related Work}
\vspace{-2mm}

%
\inlinesection{3D Generation.}
The SDS-based methods~\cite{poole2022dreamfusion,yi2023gaussiandreamer,tang2023dreamgaussian,shi2023mvdream,wang2024prolificdreamer,li2023sweetdreamer,weng2023consistent123,pan2023enhancing,chen2023text,sun2023dreamcraft3d,sargent2023zeronvs,EnVision2023luciddreamer,zhou2023dreampropeller,guo2023stabledreamer} propose to distill priors from the 2D generative model via SDS loss to optimize the 3D content from text or image.
Another line of works~\citep{hong2023lrm,jiang2022LEAP,wang2023pf,zou2023triplane,wei2024meshlrm,tochilkin2024triposr} directly predict the 3D model of an static object via a large reconstruction model.
Other approaches~\citep{liu2023zero,liu2023syncdreamer,long2023wonder3d,voleti2024sv3d,ye2023consistent,karnewar2023holofusion,instant3d2023,shi2023toss,shi2023zero123++,wang2023imagedream,liu2024one,liu2023one2} first generate dense consistent multi-view images, which are used for 3D content reconstruction.
We follow this strategy, but generate consistent multi-view videos instead of images and then reconstruct the 4D object.

\inlinesection{Video Generation.}
Recent advancements in video generation models~\citep{ho2022video,voleti2022mcvd,blattmann2023align,blattmann2023stable,he2022lvdm,singer2022make,guo2023animatediff} have shown very impressive performance with consistent geometry and realistic motions. 
Video generation models demonstrate strong generalization capabilities due to their training on large-scale image and video datasets, which are more readily available than 3D or 4D data.
In this work, we adapt the pre-trained video generation model to generate multi-view videos for 4D generation.

\begin{figure*}[t!]
    \centering
    \includegraphics[width=.95\linewidth]{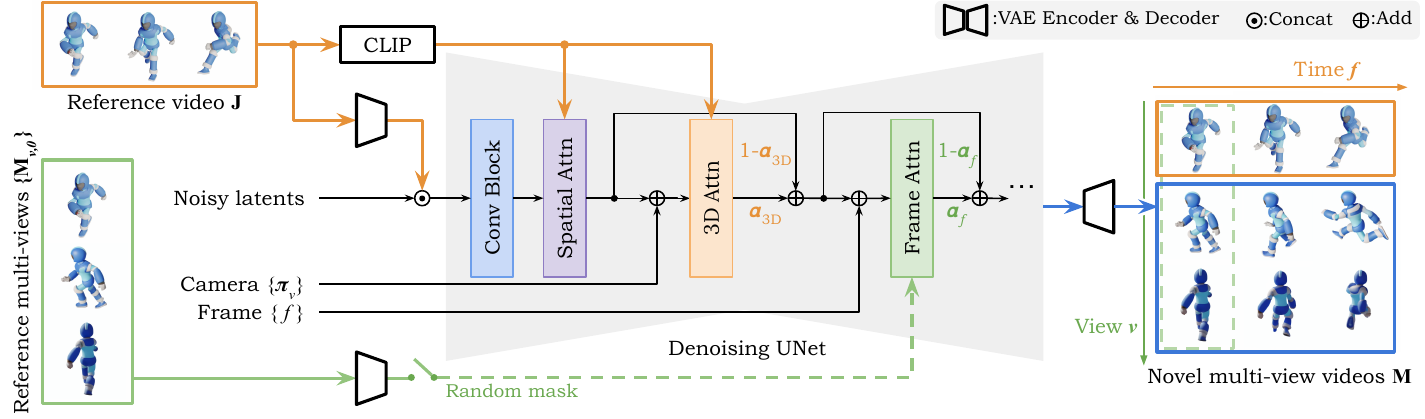}
    \vspace{-1mm}
    \caption{
        {\bf Stable Video 4D 2.0 (\name) network architecture.} Our model is similar to SV4D~\cite{xie2024sv4d} with several key differences: 1) we randomly mask the reference multi-view latents for cross-attention conditioning, allowing the model to generate multi-view videos without dependency of a separate multi-view diffusion model; 2) we replace view attention with 3D attention and condition it on the camera poses relative to the input view, making it more robust to arbitrary and sparse novel views; 3) we design an $\alpha$-blending strategy for both 3D and frame attention layers to merge spatial and temporal information effectively while preserving priors from multi-view and video diffusion models, which also enables joint training on 3D and 4D data.
    }
    \vspace{-1mm}
\label{fig:network}
\end{figure*}

\inlinesection{4D Generation.}
Recent works generate 4D content by separately utilizing pre-trained video generative model and multi-view generative model, either through SDS-based optimization~\citep{singer2023text4d,bah20244dfy,ling2023alignyourgaussians,ren2023dreamgaussian4d,zhao2023animate124,jiang2023consistent4d,zeng2024stag4d,yin20234dgen,chen2024ct4d,rahamim2024bringing,geng2024birth,yu20244real} or some inference-only pipelines~\citep{yang2024diffusion,pan2024fast,sun2024eg4d,sun2024dimensionx}. 
However, SDS-based methods tend to take hours to generate 4D content, while inference-only pipelines struggle to maintain satisfactory spatio-temporal consistency. 
A different line of approaches~\cite{liang2024diffusion4d,zhang20244diffusion,li2024vividzoo,ren2024l4gm,xie2024sv4d,yang2024diffusion,zhao2024genxd,wang20244real-video,wu2024cat4d,jiang2024animate3d,zhu2025ar4d,bai2024syncammaster} train a unified 4D generative model using synthetic 4D data. 
For instance, SV4D~\cite{xie2024sv4d} incorporates view and frame attention layers to maintain spatio-temporal consistency. 
However, these works often produce blurry details and poor spatio-temporal consistency, particularly when handling large object motion and self-occlusion.
Building upon the SV4D framework, we make several key improvements to the network architecture, training strategy, data curation, and 4D optimization.
These enhance the output quality in terms of detail preservation and spatio-temporal consistency, while better handling self-occlusion, large motion, and longer video generation.

\vspace{-2mm}
\section{Multi-view Video Synthesis via SV4D 2.0}
\vspace{-2mm}

SV4D 2.0 generates dynamic 3D content by first synthesizing multiple novel-view videos given a monocular video as input then optimizing a 4D representation via photogrammetry. 
Similar to SV4D~\cite{xie2024sv4d}, we leverage the temporal prior in a video diffusion model as well as the 3D prior in a multi-view diffusion model, while introducing key improvements in both multi-view video synthesis and 4D optimization.


\subsection{Problem Setting}
\vspace{-1mm}

Given a monocular video ${\bf J} \in \mathbb{R}^{F \Times D}$ of a dynamic object as input, where $F$ is the number of frames and $D \coloneq 3 \Times H \Times W$ is the merged image dimension, our goal is to generate novel multi-view videos ${\bf M} \in \mathbb{R}^{V \Times F \Times D}$ at $V$ views that are consistent spatially (i.e. $V$'s dimension) and temporally (i.e. $F$'s dimension).
The generated novel views must follow a conditioning camera trajectory $\bm{\pi} \in \mathbb{R}^{V \Times 2}$ specified simply by a sequence of elevation $e$ and azimuth angles $a$ relative to the input view of the monocular video.
%
%
Prior work SV4D~\cite{xie2024sv4d} generates novel multi-view videos ${\bf M}$, by learning the conditional distribution $p({\bf M}\ |\ \{ {\bf M}_{v,0} \}, {\bf J}, \bm{\pi})$ parameterized by a 4D diffusion model, where $\{ {\bf M}_{v,0} \}$ denotes the multi-view images of the first frame obtained using SV3D~\cite{voleti2024sv3d}. In contrast, SV4D 2.0 directly learns $p({\bf M}\ |\ {\bf J}, \bm{\pi})$ without additionally conditioning on the output of SV3D.

\subsection{Network Architecture}
\vspace{-1mm}

The model architecture of SV4D 2.0, as shown in \cref{fig:network}, consists of a denoising UNet, where each layer consists of a residual block with Conv3D layers, and three transformer blocks with spatial, 3D, and frame attention respectively. To leverage the rich motion and multi-view priors learned from large-scale video and 3D datasets, the weights of the frame attention layers are initialized from the corresponding layers in SVD~\citep{blattmann2023stable}, and the rest from SV3D$_p$~\citep{voleti2024sv3d}.
While our architecture shares similarities with those of SV3D~\cite{voleti2024sv3d} and SV4D~\cite{xie2024sv4d}, we make important changes in SV4D 2.0 that lead to enhanced NVVS quality.


\inlinesection{3D Attention for Sparse-view Synthesis.}
To better handle correspondence between sparse novel views i.e. views with significant distance between them, we utilize 3D attention layer~\cite{shi2023mvdream,wang2023imagedream}, while SV3D~\cite{voleti2024sv3d} and SV4D~\cite{xie2024sv4d} use view attention layers.
Let $\mathbf{L}$ denote the latent features before being fed to each attention layer $\gamma_{(\cdot)} (\cdot)$. The view attention module $\gamma_{v}$ in~\cite{voleti2024sv3d,xie2024sv4d} learns 3D consistency by attending to the same spatial location across views by first reshaping the latents as:
\vspace{-1mm}
\begin{equation}
\gamma_{v} \Big( \text{reshape}\Big(\mathbf{L}, ~ F~V~H~W~C \rightarrow  (FHW)~\underline{V}~C\Big) \Big),
\end{equation}
\vspace{-1mm}
where $C$ is the number of latent channels.
However, the same spatial location in a neighboring view need not correspond to the same 3D region, especially when generating sparse novel views.
Thus, we adopt 3D attention $\gamma_{3D}$ by attending to merged view and spatial dimensions as:
\vspace{-1mm}
\begin{align}
&  \gamma_{3D} \Big(~\text{reshape}\big(\mathbf{L}, ~ F~V~H~W~C \rightarrow  F~\underline{(VHW)}~C\big) ~\Big).
\end{align}
\vspace{-1mm}
We show in our ablation study (\cref{tab:nvs-objaverse}) that this considerably improves 3D consistency across sparse novel views, which also allows us to generate longer videos under the same compute/memory constraints.

\inlinesection{Blending 3D and Temporal Information.}
To combine the spatial and temporal information with flexible control, we use an $\alpha$-blending mechanism. 
Specifically, we introduce learnable blending weights $\alpha_{3D}$ and $\alpha_{f}$ for the 3D and frame attention block respectively, to merge the attended latents with their skip-connected versions.
Compared to SV4D~\cite{xie2024sv4d} which uses the same blending weights after multi-view and frame attention (\ie, $\alpha_{3D}=\alpha_f$), this design can more effectively preserve the 3D prior in a multi-view diffusion model and temporal prior in a video diffusion model, while learning the joint spatio-temporal consistency from 4D data.
Furthermore, by setting $\alpha_{f}=0$ to bypass the frame attention layers, we can train the model with more abundant 3D data for better generalization.
In 4D training, we initialize $\alpha_{f}$ to a near-zero value to speed up convergence by allowing the model to maximally reuse the priors in 3D pre-training.

\inlinesection{Camera and Frame Index Conditioning.}
To condition on the camera and frame index, we add the learned embeddings of camera trajectory $\{\pi_v\}$ and frames indices $\{f\}$ to the latents before 3D and frame attention layers, respectively.
In contrast to SV4D~\cite{xie2024sv4d} which uses view indices $\{v\}$ to condition view attention, camera conditioning allows the model to flexibly generate novel views following an arbitrary camera trajectory.
Note that our conditioning camera trajectory is relative to the input view, whereas SV4D takes in world-space camera poses and thus requires additional pose estimation of the input video during inference.

\inlinesection{Optional Conditioning of Reference Multi-views.}
We additionally make the dependency on reference multi-views ${\bf M}_{v,0}$ optional, by randomly masking their latents for cross-attention conditioning during training.
Although the reference multi-views provide rich 3D guidance for NVVS, it requires either ground-truth 3D renders or synthesized novel views via a multi-view diffusion model like SV3D~\cite{voleti2024sv3d} during inference.
We remove this requirement to make inference more practically feasible. More importantly, it encourages the model to focus on temporal correspondence and motion information in all input frames (instead of the first frame only), allowing NVVS that is consistent with the input video.

Note that many existing 4D generation methods~\cite{yang2024diffusion,xie2024sv4d,yin20234dgen,sun2024eg4d,ren2024l4gm,yu20244real,wang20244real-video} generate such reference multi-view images conditioned on a single anchor frame (typically frame 0), based on the underlying assumption that the appearance of a 3D object from novel views (i.e. $\{{\bf M}_{v,0}\}$) is conditionally independent of its appearance in motion from a single view (i.e. $\{{\bf M}_{0,f}\}$).
We argue that this assumption is false for general articulated objects, and demonstrate that our model is more robust to self-occlusions in a single frame as we generate the spatio-temporal images jointly. 
We also show several examples in the supplemental material where the reference multi-views conflict with the input videos and cause blurry artifacts. 

\inlinesection{Inference Sampling.}
During inference, we sample multi-view videos of $V=4$ novel views and $F=12$ frames from the learned distribution $p({\bf M}\ |\ {\bf J}, \bm{\pi})$ without the reference multi-view condition of ${\bf M}_{v,0}$.
To extend the videos while maintaining temporal consistency, we use the generated multi-views of the last generated frame as multi-view conditioning and sample longer videos from the distribution $p({\bf M}\ |\ {\bf M}_{v,0}, {\bf J}, \bm{\pi})$.
Note that this flexible inference sampling is only enabled by the proposed random masking during training.
We linearly increase the scale of classifier-free guidance (CFG) in both view and time axes when deviating from the input view and first timestamp.

\begin{figure*}[ht!]
    \centering
    \includegraphics[width=.95\linewidth]{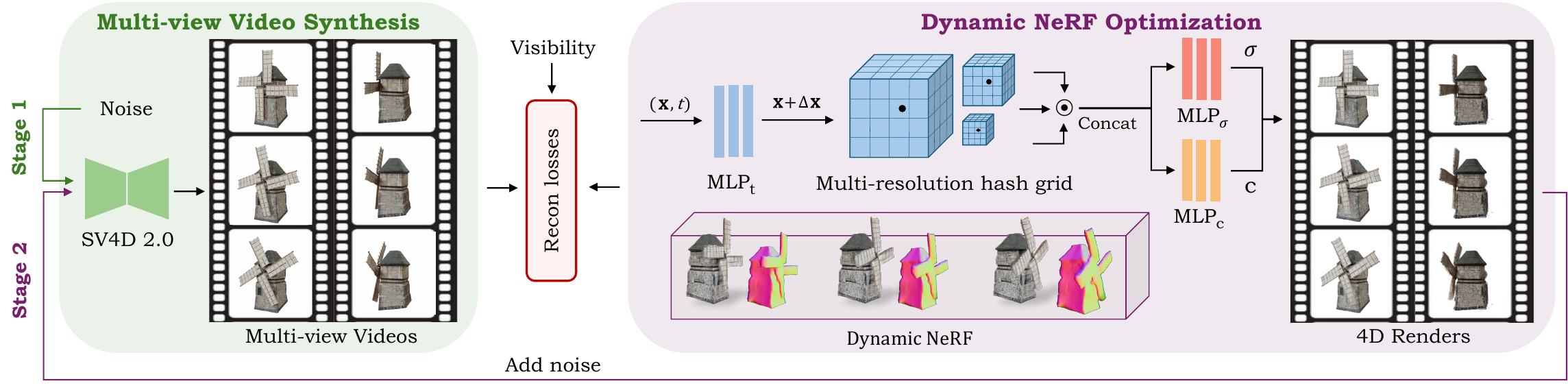}
    \vspace{-1mm}
    \caption{
        {\bf 4D optimization overview.} In the first stage, we use the initial synthesized multi-view videos as pseudo ground-truths to optimization a dynamic NeRF. To handle the 3D inconsistency and pose misalignment in novel view synthesis, we propose a second-stage refinement by noising and denoising the renders of dynamic NeRF as enhanced (3D consistent) photogrammetry targets. We also propose a visibility weighting scheme for the reconstruction losses to mitigate inconsistent texture across views.
    }
    \vspace{-1mm}
\label{fig:4d_optimization}
\end{figure*}

\subsection{Data Curation}
\vspace{-1mm}

We curate a 4D dataset, ObjaverseDy++, to train our model, which consists of renders of dynamic objects from the Objaverse~\cite{deitke2023objaverse} and ObjaverseXL datasets~\cite{deitke2023objaversexl}.
It is a quality-enhanced version of the ObjaverseDy dataset in~\cite{xie2024sv4d}.
Particularly, we prevent objects from moving off-center to facilitate the learning of temporal consistency. 
To this end, we propose to disentangle the global and local motion of each dynamic 3D object, by 1) calculating the mean temporal offsets of each surface point, 2) identifying the static regions via simple thresholding, 3) computing the average offset within a most static 3D bounding box, then 4) subtracting it globally.
We observe that such global motion calibration before rendering 4D objects leads to much higher spatio-temporal consistency in the multi-view video synthesis.
Moreover, we filter out objects with small motion or inconsistent scaling and reduce baked-in lighting effect that sometimes causes dark back views.
Lastly, we render denser spatial views and longer animation sequences for each object.
More data curation details are shown in supplemental material.

\subsection{Progressive 3D-to-4D Training}
\vspace{-1mm}

%
While most 4D generation methods~\cite{yang2024diffusion,liang2024diffusion4d,ren2024l4gm,li2024vividzoo} directly finetune a modified diffusion model on 4D data, we argue that it is non-optimal to adapt to the new network architecture and bridge the 3D-4D or video-4D domain gap.
Instead, we train the SV4D 2.0 model in a progressive 3D-to-4D manner, facilitating 4D learning while preserving rich 3D priors.
Concretely, we first train the model on the Objaverse dataset (static 3D) in~\cite{voleti2024sv3d} while bypassing the frame attention layers by setting $\alpha_f = 0$. 
Then, we unfreeze the frame attention layers as well as $\alpha_f$ 
and finetune the model on the ObjaverseDy++ dataset to boost the learning of object motion.
Although our model architecture allows joint 3D and 4D training from scratch, we show in our ablation study (see supplemental material) that this progressive learning reaches the best tradeoff between training efficiency and model performance considering the imbalanced size of 3D and 4D datasets.
The training objective of SV4D 2.0 can be expressed as:
    $\mathbb{E}_{z_n, n, {\bf J}, \bm{\pi}, \epsilon} = \Big[ \| \epsilon - \epsilon_\theta (z_n, n, {\bf J}, \bm{\pi}) \|^2_2 \Big]$,
where $\epsilon_\theta$ is the denoising network and $z_n$ is the noisy latents at denoising timestep $n$ with added noise $\epsilon$. 

\begin{figure*}[ht!]
    \centering
    \includegraphics[width=.95\linewidth]{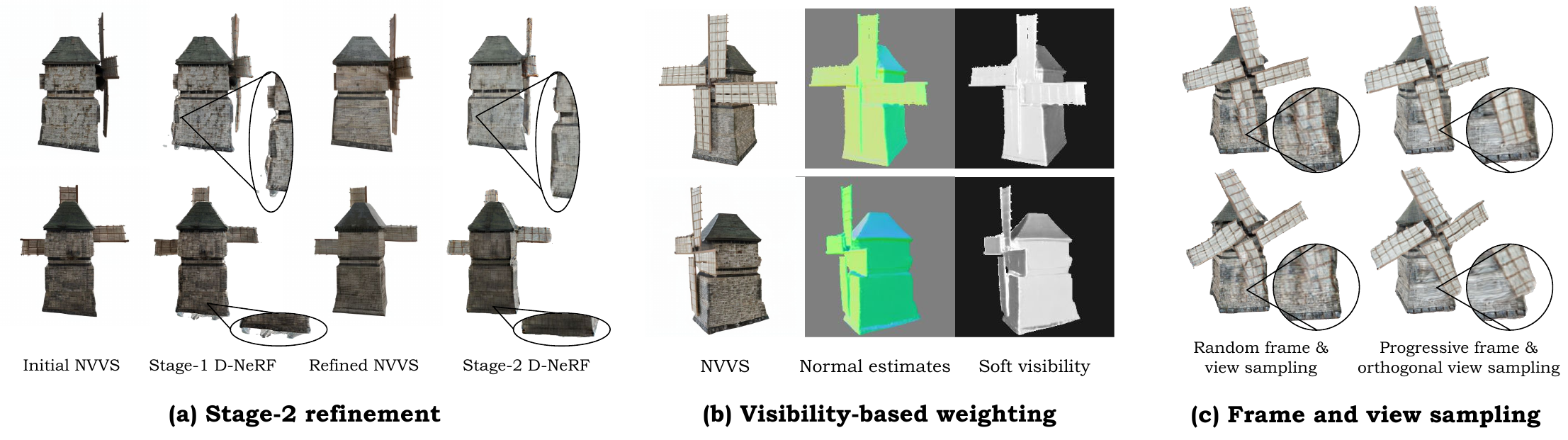}
    \vspace{-2mm}
    \caption{
        {\bf Detailed analyses of our 4D optimization strategies.} (a) Our Stage-2 refinement can effectively reduce the artifacts in dynamic NeRF caused by inconsistent novel-view video synthesis (NVVS). (b) We compute soft visibility maps as view-dependent loss weights based on surface normal estimates to further mitigate texture inconsistency. (c) The proposed progressive frame and orthogonal view sampling are shown to facilitate the learning of temporal deformation and capture better details in motion.
    }
    \vspace{-1mm}
\label{fig:4d_ablation}
\end{figure*}

\vspace{-2mm}
\section{4D Optimization from Multi-view Videos}
\vspace{-2mm}

\inlinesection{4D Representation and Photogrammetry-based Optimization.}
%
As shown in \cref{fig:4d_optimization}, by using the multi-view videos generated by SV4D 2.0 as pseudo ground-truths, we adopt photogrammetry-based optimization to learn a dynamic NeRF representation for each 4D asset.
Formally, the implicit NeRF representation can be expressed as $\Psi: (\mathbf{x}, \mathbf{t}) \mapsto (\sigma, \mathbf{c})$, mapping each sampled 3D point $\mathbf{x} = (x,y,z)$ and time embedding $\mathbf{t}$ to the corresponding density $\sigma \in \mathbb{R}_+$ and color $\mathbf{c} \in \mathbb{R}^3_+$, which can then be used to render images/videos by densely sampling points along camera rays and timestamps.
Similar to D-NeRF~\citep{pumarola2021d} and SV4D~\cite{xie2024sv4d}, we compose a 4D object with a canonical NeRF and a temporal deformation field, where the former is parametrized by a multi-resolution hash grid~\citep{mueller2022instant} with density and color MLPs (MLP$_\mathrm{\sigma}$ and MLP$_\mathrm{c}$), and the later by a deformation MLP (MLP$_\mathrm{t}$).

%
Instead of relying on the cumbersome SDS losses from image/video diffusion models, we follow SV4D~\cite{xie2024sv4d} to optimize a dynamic NeRF via photogrammetry.
The main reconstruction losses include the pixel-level mean squared error $\mathcal{L}_\text{mse} = \lVert \bm{M}-\bm{\hat{M}} \rVert^2$, perceptual LPIPS~\citep{zhang2018unreasonable} loss $\mathcal{L}_\text{lpips}$, and mask loss $\mathcal{L}_\text{mask} = \lVert \bm{S} - \bm{\hat{S}} \rVert^2$, where $\bm{S}$, $\bm{\hat{S}}$ are the predicted and ground-truth silhouettes.
We learn the 4D representation in a static-to-dynamic manner to improve training efficiency and stability.
That is, we first optimize the canonical NeRF only on the multi-view images of the first frame, then unfreeze MLP$_\mathrm{t}$ while sampling multiple views and frames for joint optimization.


\inlinesection{Improving 3D Consistency and Pose Alignment.}
Although SV4D~\cite{xie2024sv4d} and other prior methods~\cite{yang2024diffusion,sun2024eg4d,liang2024diffusion4d} demonstrate that the such photogrammetry-based optimization can efficiently create a 4D object without SDS losses, we observe that the slight 3D inconsistency or pose misalignment (generated view not aligned with input camera pose) in multi-view video synthesis can greatly affect the 4D output quality, resulting in undesirable artifacts or blurry details.
To address this, we propose a 2-stage optimization framework and weighted reconstruction loss based on soft visibility estimation, as shown in~\cref{fig:4d_optimization}.

In the first stage, we sample multi-view videos from noise as pseudo ground-truths and optimize a dynamic NeRF via the aforementioned reconstruction losses.
In the second stage, we refine the pseudo ground-truth videos by noising and denoising the rendered views and frames of the stage-1 dynamic NeRF.
Since these videos are rendered and denoised from an optimized 4D representation, they are by nature more consistent and pose-aligned in 3D compared to the initial video synthesis from noise, thus providing a better reconstruction target.
Note that such iterative refinement can be applied multiple times~\cite{yu2024viewcrafter} and at each denoising step~\cite{tang2024cycle3d,chen20243d} with a tradeoff for optimization speed.
In practice, we find that adding a second-stage optimization already achieves noticeably better output quality in terms of 3D coherence and detail sharpness, as shown in~\cref{fig:4d_ablation} (a).

To better deal with the inconsistent texture across views, we also propose a loss weighting scheme by treating visibility information as confidence for each generated view.
Concretely, for each foreground pixel $\bm{p}$ in view $v$, we estimate its soft visibility $\bm{W_p}$ based on the dot product between camera ray direction $\bm{v}$ (calculated from camera pose $\bm{\pi_v}$) and surface normal $\bm{n}$ (estimated from Omnidata~\cite{Eftekhar2021omnidata}) as: $\bm{W_p} = \bm{v} \cdot \bm{n}.$
Since higher values of $\bm{W_p}$ indicate better visibility of the surface from the camera, we clip the range of $\bm{W}$ to $(0,1)$ and use it as confidence weights for the MSE loss: $\mathcal{L}_\text{mse} = \lVert \bm{W} (\bm{M}-\bm{\hat{M}}) \rVert^2$.
We show examples of the visibility maps in~\cref{fig:4d_ablation} (b).

\inlinesection{View and Frame Sampling for Stable 4D Optimization.}
Due to memory limitation, SV4D~\cite{xie2024sv4d} randomly samples 4 views and 4 frames during 4D optimization.
To further stabilize training, we propose orthogonal view sampling and progressive frame sampling to stabilize training.
Specifically, we sample near-orthogonal views (\eg, azimuth degree $a=0, 90, 180, 270$) to minimize free deformation in unseen regions while progressively increasing the temporal window (range of frame indices) for frame sampling.
We show in~\cref{fig:4d_ablation} (c) that this sampling strategy effectively improves the learning of temporal deformation field in presence of large motion.

\vspace{-2mm}
\section{Experiments}
\vspace{-2mm}

\inlinesection{Datasets.}
We evaluate the results of NVVS and 4D optimization on two synthetic datasets: ObjaverseDy~\cite{deitke2023objaverse,xie2024sv4d} and Consistent4D~\citep{deitke2023objaverse}.
For ObjaverseDy, we use the same validation objects as~\cite{xie2024sv4d} and render 21 frames $\times$ 8 views per object for evaluation.
We run all baseline models on our validation set using their official code, conditioned on a near-front view video for each object.
Consistent4D dataset contains multi-view renders of several dynamic objects from Objaverse~\citep{deitke2023objaverse}. 
We used the same input video and evaluate on the same 4 novel views as~\cite{jiang2023consistent4d,xie2024sv4d}.
Additionally, we use monocular videos from the real-world video dataset DAVIS~\citep{Perazzi_CVPR_2016, Pont-Tuset_arXiv_2017, Caelles_arXiv_2019} for the visual comparison and user study.

\inlinesection{Metrics.}
We evaluate the visual quality of our video and 4D outputs using common image metrics like LPIPS~\citep{zhang2018unreasonable}, CLIP-S, PSNR, SSIM, and MSE.
Following~\cite{xie2024sv4d}, we also evaluate spatio-temporal consistency by calculating FVD~\citep{unterthiner2018towards} in different ways:
\textit{FVD-F}: over frames at a fixed view.
\textit{FVD-V}: over views at a fixed frame.
\textit{FVD-Diag}: over the diagonal images of the image matrix.
\textit{FV4D}: over all images by scanning them in a bidirectional raster (zig-zag) order.

\inlinesection{Baselines.}
For \textit{NVVS}, we compare \name with several recent methods capable of generating multiple novel-view videos from a monocular video, including
\nameold~\cite{xie2024sv4d},
per-frame SV3D~\citep{voleti2024sv3d}, Diffusion$^2$~\citep{yang2024diffusion},
4Diffusion~\citep{zhang20244diffusion}.
For \textit{4D generation}, we compare \name with other baselines that can generate 4D representations, including 
SV4D~\cite{xie2024sv4d},
Consistent4D~\citep{jiang2023consistent4d}, STAG4D~\citep{zeng2024stag4d}, 
4Diffusion~\citep{zhang20244diffusion},
DreamGaussian4D (DG4D)~\citep{ren2023dreamgaussian4d}, GaussianFlow~\citep{gao2024gaussianflow}, 4DGen~\citep{yin20234dgen}, Efficient4D~\citep{pan2024fast}, L4GM~\cite{ren2024l4gm}.
We omit other prior methods since they either cannot perform video-to-4D generation~\cite{jiang2024animate3d,rahamim2024bringing,geng2024birth} or do not release the code or model weights at the time of submission~\cite{li2024vividzoo,zhao2024genxd,yu20244real,wang20244real-video,wu2024cat4d,sun2024dimensionx,bai2024syncammaster,liang2024diffusion4d,geng2024birth}.

\begin{figure*}[t!]
    \centering
    \includegraphics[width=.95\linewidth]{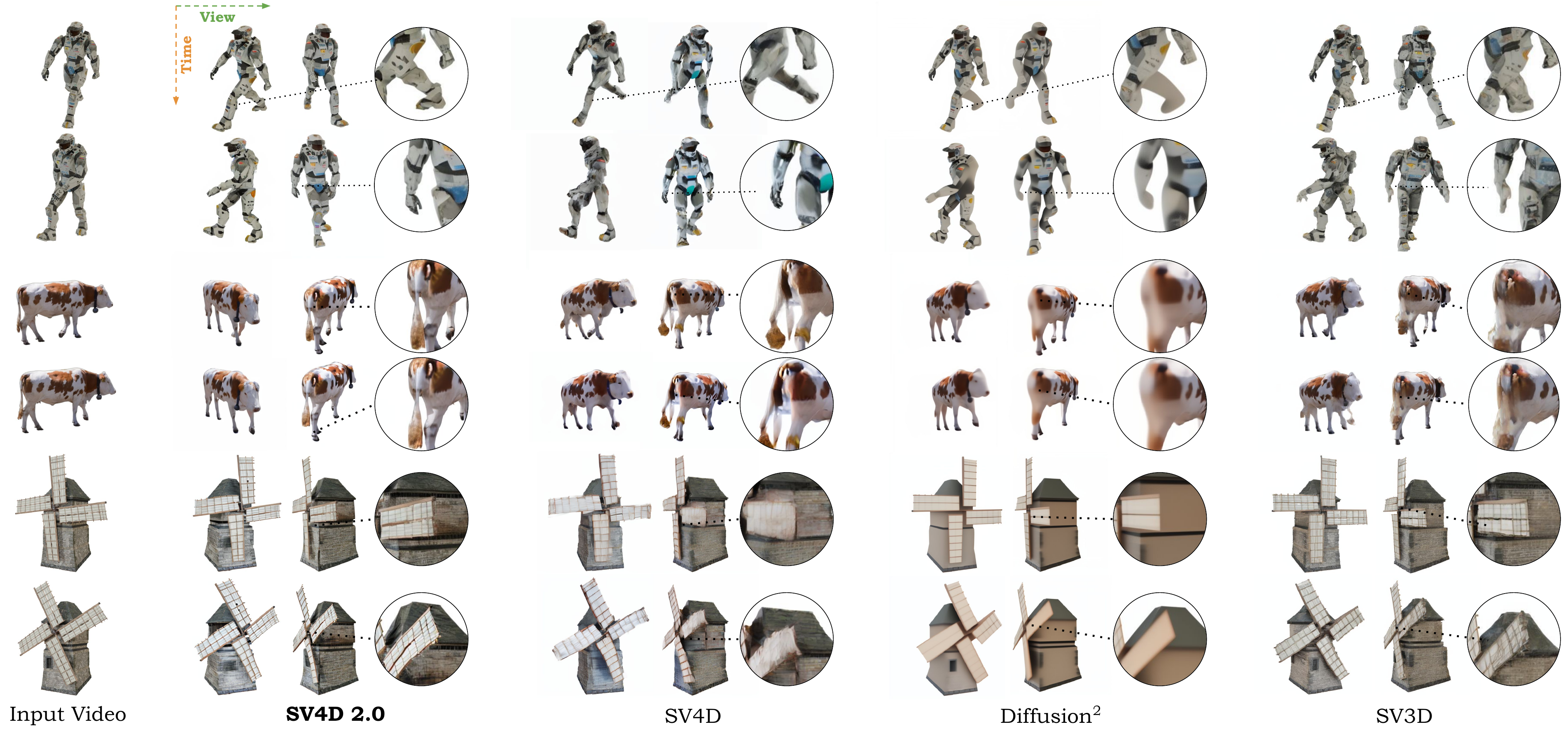}
    \vspace{-2mm}
    \caption{
        \textbf{Visual Comparison of Novel View Video Synthesis Results.}
        We show two frames in the input videos and two novel-view results of the corresponding frames. Compared to the baseline methods, \name outputs contain geometry and texture details that are more faithful to the input video and consistent across frames.
        \textit{
        We also refer reviewers to the Supplemental Material for the video comparison.
        }
    }
    \vspace{-1mm}
\label{fig:nvs_results}
\end{figure*}
\begin{figure*}[t!]
    \centering
    \includegraphics[width=.95\linewidth]{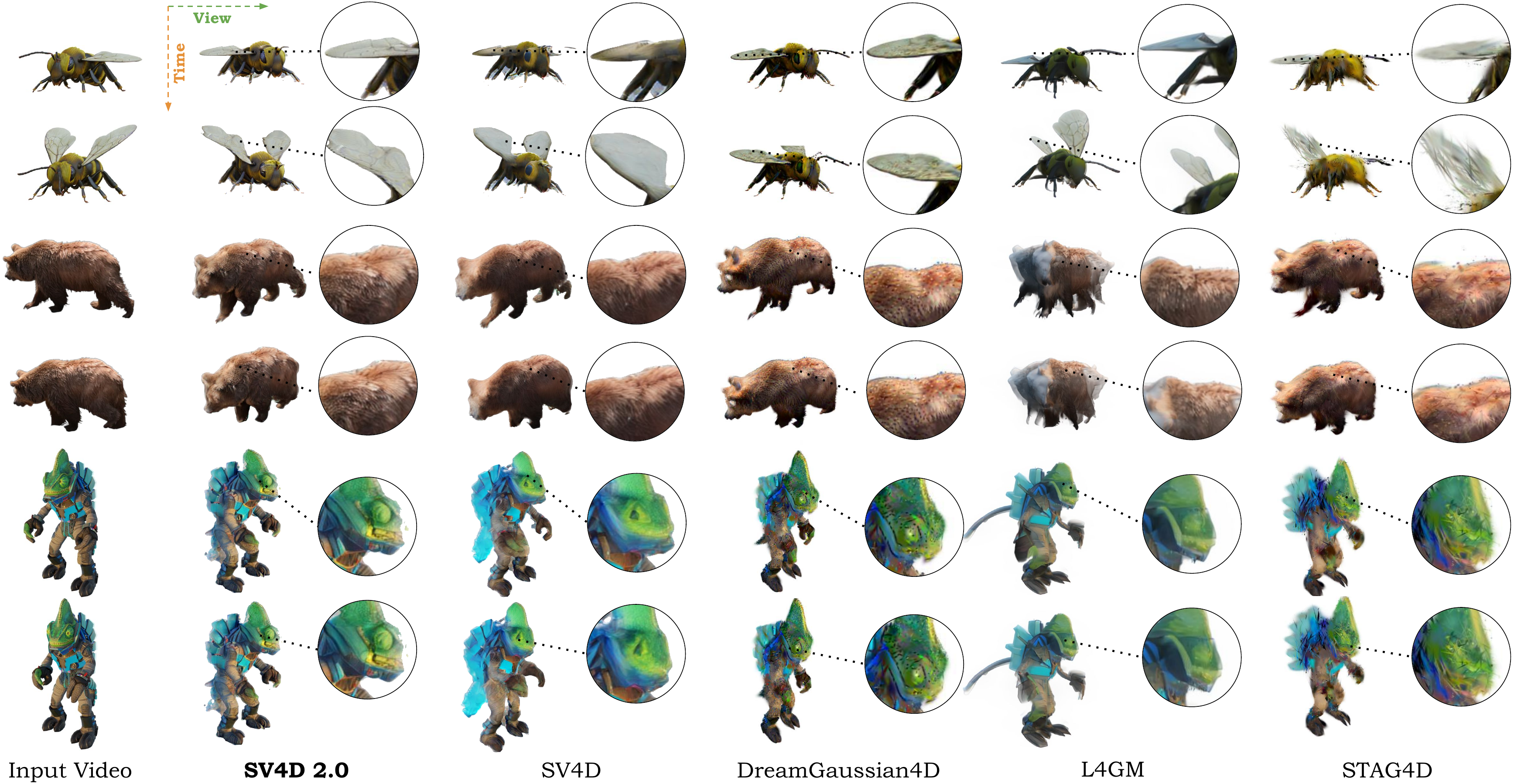}
    \vspace{-2mm}
    \caption{
        \textbf{Visual Comparison of 4D Optimization Results.}
        We show two frames in the input videos and render a novel view of the corresponding frames. By leveraging the detailed and consistent multi-view videos generated by \name, we can optimize higher-quality 4D assets compared to prior works. 
        In particular, our results are more detailed, consistent, and faithful to the input videos.
    }
\label{fig:opt_results}
\vspace{-1mm}
\end{figure*}

\subsection{Visual Comparison}
\vspace{-1mm}

In~\cref{fig:nvs_results,fig:opt_results}, we show visual comparison against prior methods on multi-view video synthesis and 4D generation, respectively.
We observe that applying SV3D frame-by-frame can preserve geometry and texture details across novel views but it fails to maintain temporal consistency.
Diffusion$^2$ tends to produce over-smoothed results by sacrificing details for temporal coherence.
\nameold can generate novel-view videos with decent spatio-temporal consistency, but it often produces blurry details during large motion. 
By comparison, \name outputs are higher-quality in terms of details, sharpness and consistency across views and frames.
In 4D generation (\cref{fig:opt_results}), STAG4D often produce artifacts caused by its Gaussian representation and inconsistent novel views. 
DG4D generates finer details but suffers from over-saturated texture, temporal flickering, and inaccurate geometry, due to SDS-based optimization.
L4GM does not generalize well on real-world data (no video prior like ours) and struggles with videos at non-zero elevation (training data primarily at 0° elevation).
\nameold improves with cleaner texture and geometry, though it fails to capture thin geometric structures or texture details.
In contrast, \name can generate high-fidelity 4D assets with realistic geometry, texture, and motion.
We also show ablative results in~\cref{fig:4d_ablation} to justify the effectiveness of our 4D optimization strategies.


\definecolor{best}{rgb}{0.996, 0.769, 0.3098}
\newcommand{\C}{\cellcolor{best}}

\definecolor{second}{rgb}{1, 0.969, 0.733}
\newcommand{\CC}{\cellcolor{second}}

\begin{table*}[t!]
\caption{
\textbf{Evaluation of NVVS on the ObjaverseDy dataset.}
\name achieves better visual quality (\textit{LPIPS}, \textit{CLIP-S}, \textit{PSNR}, \textit{SSIM}, \textit{MSE}), video frame consistency (\textit{FVD-F}), multi-view consistency (\textit{FVD-V}), and spatio-temporal consistency (\textit{FVD-Diag} and \textit{FV4D}).
}
\vspace{-.75em}
\label{tab:nvs-objaverse}
\centering
\resizebox{0.96\textwidth}{!}{
\begin{tabular}{ l c c c c c c c c c c }
\toprule 
 Model & LPIPS$\downarrow$ & CLIP-S$\uparrow$ & PSNR$\uparrow$ & SSIM$\uparrow$ & MSE$\downarrow$ & FVD-F$\downarrow$ & FVD-V$\downarrow$ & FVD-Diag$\downarrow$ & FV4D$\downarrow$ \\
 \midrule
 SV3D~\citep{voleti2024sv3d} & 0.131 & 0.922 & 17.60 & 0.873 & 0.0244 & 799.73 & 397.37 & 531.85 & 754.09\\
 Diffusion$^2$~\citep{yang2024diffusion} & 0.133 & 0.892 & 18.08 & \C \textbf{0.886} & 0.0225 & 987.19 & 477.66 & 789.12 & 1020.15\\
 \nameold~\cite{xie2024sv4d} & 0.122 & 0.919 & 17.83 & 0.881 & 0.0217 & 647.97 & 427.97 & 564.24 & 458.34\\
 \midrule
  \name w/o 3D attention  & 0.115 & \CC 0.934 & 17.89 & 0.878 & 0.0214 & \CC 548.14 & 368.68 & \CC 463.59 & \CC 283.10 \\
  \name w/o improved data  & \CC 0.112 & 0.931 & \CC 18.19 & \CC 0.884 & \CC 0.0192 & 576.66 & \CC 344.68 & 467.57 & 316.91 \\
  \name (Ours) & \C \textbf{0.105} & \C \textbf{0.939} & \C \textbf{18.67} & \C \textbf{0.886} & \C \textbf{0.0180} & \C \textbf{473.90} & \C \textbf{272.85} & \C \textbf{426.92} & \C \textbf{256.84} \\
\bottomrule
\end{tabular}
}
\end{table*}
\begin{table*}[t!]
\caption{
\textbf{Evaluation of 4D outputs on the ObjaverseDy dataset.} 
\name consistently outperforms baselines in all metrics.
}
\vspace{-.75em}
\label{tab:4d-objaverse}
\centering
\resizebox{0.90\textwidth}{!}{
\begin{tabular}{ l c c c c c c c c c }
\toprule 
 Model & LPIPS$\downarrow$ & CLIP-S$\uparrow$ & PSNR$\uparrow$ & SSIM$\uparrow$ & MSE$\downarrow$ & FVD-F$\downarrow$ & FVD-V$\downarrow$ & FVD-Diag$\downarrow$ & FV4D$\downarrow$\\
 \midrule
 Consistent4D~\citep{jiang2023consistent4d} & 0.148 & 0.899 & 16.44 & 0.866 & 0.0297 & 781.38 & 510.04 & 782.79 & \CC 658.31\\
 STAG4D~\citep{zeng2024stag4d} & 0.155 & 0.868 & 16.73 & 0.867 & 0.0260 & 848.83 & 539.96 & 709.52 & 833.08 \\
 DG4D~\citep{ren2023dreamgaussian4d} & 0.156 & 0.874 & 16.02 & 0.860 & 0.0293 & 826.72 & 543.29 & 761.58 & 741.99 \\
 L4GM~\cite{ren2024l4gm} & 0.146 & \CC 0.902 & 17.65 & 0.877 & 0.0227 & 805.87 & 537.46 & 782.92 & 666.48 \\ \nameold~\cite{xie2024sv4d} & \CC 0.122 & \CC 0.902 & \CC 18.47 & \CC 0.884 & \CC 0.0189 & \CC 754.23 & \CC 436.86 & \CC 666.59 & 699.04 \\
  \midrule
 \name (Ours) & \C \textbf{0.107} & \C \textbf{0.916} & \C \textbf{19.37} & \C \textbf{0.890} & \C \textbf{0.0153} & \C \textbf{654.64} & \C \textbf{375.03} & \C \textbf{513.26} & \C \textbf{532.55} \\
\bottomrule
\end{tabular}
\vspace{-2mm}
}
\end{table*}

\subsection{Quantitative Comparison}
\vspace{-1mm}

For \textit{NVVS}, we show the quantitative comparison on the ObjaverseDy and Consistent4D datasets in~\cref{tab:nvs-objaverse,tab:nvs-consistent4d}, respectively.
The results on both datasets show a consistent performance gain by SV4D 2.0 over all baselines on all metrics.
In particular, the pixel-level metrics (PSNR, SSIM, MSE) demonstrate that SV4D 2.0 outputs have higher image-quality and align better with the ground-truths.
The perceptual-level metrics (LPIPS, CLIP-S) show that our results are semantically more faithful to the objects.
More importantly, SV4D 2.0 achieves much lower FVD metrics across frames, views, diagonal images, and full image matrix, justifying its superior spatio-temporal consistency.
We also show several ablative results (w/o 3D attention, w/o improved data curation in ObjaverseDy++) in~\cref{tab:nvs-objaverse} to validate the effectiveness of our model design. 
%
%
%
For \textit{4D generation}, we evaluate the rendered videos on the ObjaverseDy and Consistent4D datasets in~\cref{tab:4d-consistent4d,tab:4d-objaverse}.
The 4D results also show that SV4D 2.0 consistently achieves higher quality in terms of both image and video quality.
L4GM~\cite{ren2024l4gm} performs comparably to our method on Consistent4D but struggles with videos at non-zero elevation in ObjaverseDy.
We show more results and ablations in the supplemental material.

\begin{table}[t!]
    \caption{
        \textbf{Evaluation of NVVS on Consistent4D.} SV4D 2.0 achieves a consistent performance gain on image and video metrics. 
    }
    \label{tab:nvs-consistent4d}
    \vspace{-.75em}
    \resizebox{0.85\textwidth}{!}{
    \begin{tabular}{ l c c c c}
        \toprule 
         Model & LPIPS$\downarrow$ & CLIP-S$\uparrow$ & FVD-F$\downarrow$ \\
         \midrule
         SV3D~\citep{voleti2024sv3d} & \CC 0.129 & 0.925 & 989.53\\
         4Diffusion~\citep{zhang20244diffusion} & 0.164 & 0.863 & -\\
         Diffusion$^2$~\citep{yang2024diffusion} & 0.189 & 0.907 & 1205.16 \\
         \nameold~\cite{xie2024sv4d} & \CC 0.129 & \CC 0.929 & \CC 677.68\\
         \midrule
         \name (Ours) & \C \textbf{0.115} & \C \textbf{0.946} & \C \textbf{553.43} \\
        \bottomrule
    \end{tabular}
    }
    \vspace{-1.5em}
\end{table}

\subsection{User Study}
\vspace{-1mm}
To further evaluate our model performance on in-the-wild videos, we also conduct user study to compare the NVVS results of SV4D 2.0 against 3 prior methods. 
Specifically, we select 20 real-world videos from the DAVIS dataset~\cite{Caelles_arXiv_2019} with relatively steady camera and minimal truncation/occlusions. 
Following~\cite{xie2024sv4d}, we randomly choose a novel camera view for each input video and ask users to select a novel-view video (generated by 4 different methods) that ``looks more stable, realistic, and closely resembles the reference subject''.
As a result, SV4D 2.0 outputs are preferred 85.2\% over Diffusion$^2$ (1.8\%), STAG4D (5.1\%), and SV4D (7.9\%) among all participants, showing much better generalization to these challenging real-world videos.

\begin{table}[t!]
    \centering
    \caption{
            \textbf{Evaluation of 4D outputs on the Consistent4D dataset.} 
            SV4D 2.0 achieves state-of-the-art visual quality (LPIPS, CLIP-S) and temporal smoothness (FVD-F) compared to prior methods. 
    }
    \label{tab:4d-consistent4d}
    \vspace{-.75em}
    \resizebox{0.88\textwidth}{!}{
    \begin{tabular}{ l c c c c }
        \toprule 
         Model & LPIPS$\downarrow$ & CLIP-S$\uparrow$ & FVD-F$\downarrow$ \\
         \midrule
         Consistent4D~\citep{jiang2023consistent4d} & 0.160 & 0.87 & 1133.93 \\
         STAG4D~\citep{zeng2024stag4d} & 0.126 & 0.91 & 992.21\\
         4Diffusion~\citep{zhang20244diffusion} & 0.165 & 0.88 & - \\
         Efficient4D~\citep{pan2024fast} & 0.130 & 0.92 & - \\
         4DGen~\citep{yin20234dgen} & 0.140 & 0.89 & -\\ 
         DG4D~\citep{ren2023dreamgaussian4d} & 0.160 & 0.87 & -\\
         GaussianFlow~\citep{gao2024gaussianflow} & 0.140 & 0.91 & - \\
         L4GM~\cite{ren2024l4gm} & 0.120 & \C \textbf{0.94} & \C \textbf{691.87}\\
         \nameold~\cite{xie2024sv4d} & \CC 0.118 & 0.92 & 732.40\\
         \midrule
         \name (Ours) & \C \textbf{0.117} & \CC 0.93 & \CC 692.68 \\
        \bottomrule
    \end{tabular}
    }
    \vspace{-1em}
\end{table}

\vspace{-2mm}
\section{Conclusion}
\vspace{-2mm}

We present SV4D 2.0, an enhanced multi-view video diffusion model that can generate high-quality novel-view videos and 4D assets given a monocular video.
Compared to our baseline~\cite{xie2024sv4d}, SV4D 2.0 can produce longer videos, from sparser camera views, with better spatio-temporal consistency, detail sharpness, and generalizability to real-world videos.
Moreover, it does not require reference multi-views from a separate model during inference, making it more practical and robust to occlusions in anchor frame. 
We achieve these by introducing novel techniques in data curation, network architecture, progressive training, and 4D optimization.
Our extensive evaluation and user study show that SV4D 2.0 outperforms prior methods in both NVVS and 4D generation, making it a favorable 4D foundation model.


{
    \small
    \bibliographystyle{ieeenat_fullname}
    \bibliography{main}
}

\newpage
\clearpage
\setcounter{page}{1}
\maketitlesupplementary


In this supplemental document, we include the implementation details in~\cref{sec:implementation}, ablative analyses in~\cref{sec:ablation}, and additional results in~\cref{sec:additional_results}. We also attach a teaser video (\texttt{SV4D\_2.0\_video.mp4}) to summarize the SV4D 2.0 framework and show more visual results.

\section{Implementation Details}
\label{sec:implementation}

\subsection{Data curation details}

\begin{figure}[t!]
    \centering
    \includegraphics[width=\linewidth]{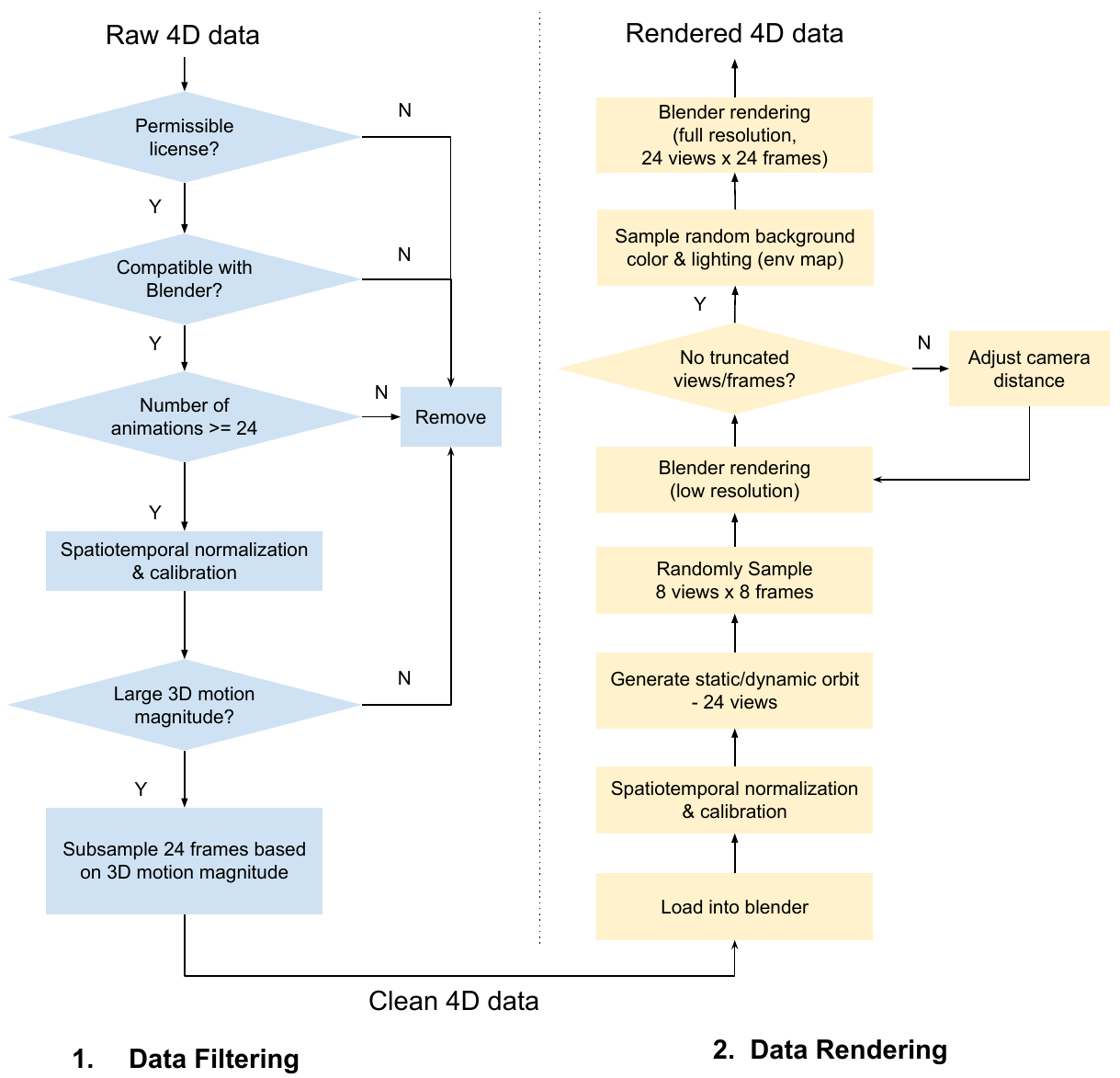}
    \caption{
        \textbf{Detailed data processing pipeline.}
    }
    \vspace{-2mm}
\label{fig:data-pipeline}
\end{figure}

\inlinesection{Dataset overview.}
We illustrate the overall data pipeline in~\cref{fig:data-pipeline}. To train SV4D 2.0, we render CC-licensed animatable 3D objects from the Objaverse~\citep{deitke2023objaverse} and ObjaverseXL datasets~\citep{deitke2023objaversexl} datasets.
Objaverse contains 44k dynamic 3D objects and ObjaverseXL includes roughly 323k in the GitHub subset.
Similar to SV4D~\cite{xie2024sv4d}, we filter out nearly half of the objects based on license, inconsistent scaling, object motion, etc.
We scale each object such that the largest world-space XYZ extent of its bounding box is 1, then use Blender's CYCLES renderer to render multiple views and video frames at 576$\times$576 resolution. For lightning, we randomly select from a set of curated HDRI environment maps. 
Following SV3D~\cite{voleti2024sv3d}, we render both static (uniform azimuth and fixed elevation) and dynamic (irregular azimuth and elevation) orbits for training.
We then encode all rendered images into the latent space using SD2.1~\citep{rombach2022high}'s VAE and CLIP~\citep{radford2021learning}.

\begin{figure}[t!]
    \centering
    \includegraphics[width=\linewidth]{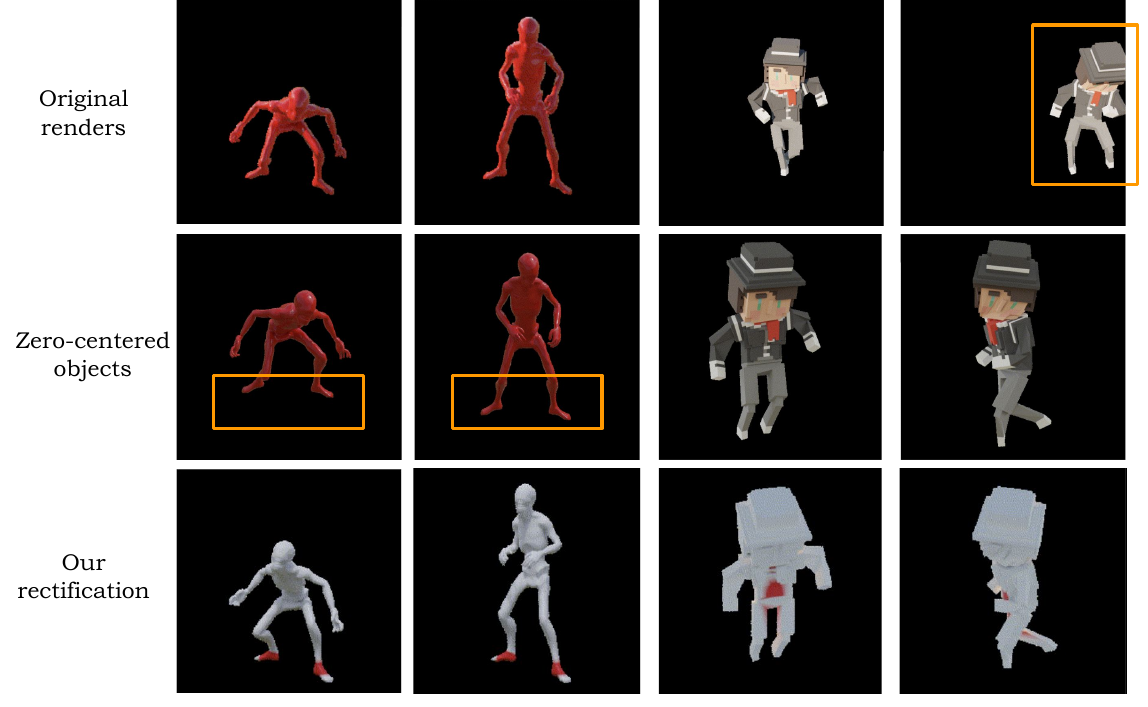}
    \caption{
        \textbf{Rectifying off-center objects.} We show in the top row (right) that some objects move off-center during motion. In the second row, we demonstrate that naively zero-positioning the center of object does not apply to some objects (left) as the object center is not always static. Instead, we propose rectification by identifying the most static region (marked in red) and disentangling the global translation from local motion, as shown in the bottom row.
    }
    \vspace{-2mm}
\label{fig:off-center}
\end{figure}

\inlinesection{Disentangling global and local motion.}
In our improved 4D dataset, ObjaverseDy++, we disentangle the global transformation from local motion of object parts to preserve temporal correspondence between frames during motion and prevent truncation or off-center cases.
As shown in~\cref{fig:off-center}, we identify the most static region by computing the mean 3D displacement of each vertex over time and computing the average offset within this region as global translation.
Then, we subtract the global translation from all vertex coordinates to fix the static region.

\inlinesection{Mitigating strong shading effect.}
Moreover, to reduce baked-in lighting effect that sometimes causes dark back views, we filter out environment maps that have strong shading effect and only sample from near-ambient maps during object rendering.
We show in~\cref{fig:results_nvs_2} (snowboarding and horse riding videos) that this effectively mitigates the issue of dark back views.

\begin{table*}[t!]
\caption{
\textbf{Ablation study of SV4D 2.0 novel-view video synthesis on the ObjaverseDy dataset~\cite{deitke2023objaverse,xie2024sv4d}.}
By showing better metrics of SV4D 2.0 compared to the ablated models, we justify our model design of 1) using VAE latents of reference multi-views to optionally condition frame attention, 2) adding 3D attention with camera embedding as condition (instead of view indices), 3) improved data curation, and 4) progressive 3D-4D training.
}
\label{tab:nvs-ablation}
\centering
\resizebox{\textwidth}{!}{
\begin{tabular}{ l c c c c c c c c c c }
\toprule 
 Model & LPIPS$\downarrow$ & CLIP-S$\uparrow$ & PSNR$\uparrow$ & SSIM$\uparrow$ & MSE$\downarrow$ & FVD-F$\downarrow$ & FVD-V$\downarrow$ & FVD-Diag$\downarrow$ & FV4D$\downarrow$ \\
 \midrule
  SV4D 2.0 w/o reference multi-views & 0.135 & 0.870 & 17.07 & 0.870 & 0.0255 & 876.07 & 410.90 & 608.93 & 710.66 \\
  SV4D 2.0 w/ CLIP of reference multi-views & 0.142 & 0.868 & 17.27 & 0.877 & 0.0244 & 1174.47 & 530.74 & 819.22 & 1327.75 \\
  SV4D 2.0 w/o 3D attention  & 0.115 & \CC 0.934 & 17.89 & 0.878 & 0.0214 & \CC 548.14 & 368.68 & \CC 463.59 & \CC 283.10 \\
  SV4D 2.0 w/ view index conditioning & 0.121 & 0.928 & 18.08 & 0.881 & 0.210 & 660.67 & 401.48 & 507.55 & 450.12 \\
  SV4D 2.0 w/o improved data  & \CC 0.112 & 0.931 & \CC 18.19 & \CC 0.884 & \CC 0.0192 & 576.66 & \CC 344.68 & 467.57 & 316.91 \\
  SV4D 2.0 w/o progressive 3D-4D training & 0.140 & 0.896 & 16.50 & 0.871 & 0.0284 & 727.60 & 451.78 & 662.22 & 475.44 \\
  \textbf{SV4D 2.0 (Ours)} & \C 0.105 & \C 0.939 & \C 18.67 & \C 0.886 & \C 0.0180 & \C 473.90 & \C 272.85 & \C 426.92 & \C 256.84 \\
\bottomrule
\end{tabular}
}
\end{table*}

\begin{table*}[t!]
\caption{
\textbf{Ablation study of 4D optimization on the ObjaverseDy dataset~\cite{deitke2023objaverse,xie2024sv4d}.} We show that our stage-2 refinement can significantly improve the 4D outputs in terms of image quality (LPIPS, PSNR, MSE, etc) and 3D consistency (FVD-V). The proposed orthogonal view and progressive frame sampling also achieves slightly higher image quality (LPIPS, CLIP-S, PSNR, etc) and temporal consistency (FVD-F) compared to random sampling.
}
\label{tab:4d-ablation}
\centering
\resizebox{0.90\textwidth}{!}{
\begin{tabular}{ l c c c c c c c c c }
\toprule 
 Model & LPIPS$\downarrow$ & CLIP-S$\uparrow$ & PSNR$\uparrow$ & SSIM$\uparrow$ & MSE$\downarrow$ & FVD-F$\downarrow$ & FVD-V$\downarrow$ & FVD-Diag$\downarrow$ & FV4D$\downarrow$\\
 \midrule
 SV4D 2.0 w/o stage-2 refinement & 0.124 & 0.910 & 17.94 & 0.876 & 0.0194 & 699.84 & 481.80 & 586.31 & 560.83 \\
 SV4D 2.0 w/ random view \& frame sampling & \CC 0.105 & \CC 0.918 & \CC 19.41 & \C 0.889 & \C 0.0153 & \CC 644.15 & \CC 378.77 & \CC 491.42 & \C 495.67 \\
 \textbf{SV4D 2.0 (Ours)} & \C 0.104 & \C 0.921 & \C 19.44 & \C 0.889 & \CC 0.0154 & \C 637.35 & \C 355.45 & \C 472.98 & \CC 511.90 \\
\bottomrule
\end{tabular}
}
\end{table*}

\subsection{SV4D 2.0 training details}
Similar to SVD~\citep{blattmann2023align}, we train our model following the widely used EMD~\citep{karras2022elucidating} scheme with $L2$ loss.
To optimize the training speed and reduce GPU VRAM, we follow SV4D to precompute the VAE latents and CLIP embeddings of all images in advance.
To form a training batch, we randomly sample 4 views and 12 frames of each object from the 24 rendered views and 24 frames.
We train SV4D 2.0 on 16 NVIDIA H100 GPUs with an effective batch size of 32 (each GPU fits 2 batches).
The model is first trained on static 3D data for 175k iterations then finetuned on 4D data for 185k iterations.

\subsection{4D optimization details}
The main supervision in our 4D optimization comes from the SV4D 2.0 NVVS results, which we sample from noise though 50 denoising steps.
For stage-2 refinement, we add noise to the initial renders at noise timestep 25 and then denoise the images as improved pseudo ground-truths.
Our optimization losses include the visibility-weighted MSE $\mathcal{L}_\text{mse} = \lVert \bm{W} (\bm{M}-\bm{\hat{M}}) \rVert^2$, LPIPS~\citep{zhang2018lpips} loss $\mathcal{L}_\text{lpips}$, mask loss $\mathcal{L}_\text{mask} = \lVert \bm{S} - \bm{\hat{S}} \rVert^2$, and a normal loss $\mathcal{L}_\text{normal}=1-(\bm{n} \cdot \bm{\bar{n}})$, where $\bm{n}$ and $\bm{\bar{n}}$ are the rendered normal and estimated pseudo ground truths from Omnidata~\citep{Eftekhar2021omnidata}, respectively.
We also adopt a smooth depth loss and bilateral normal smoothness loss proposed in SV3D~\cite{voleti2024sv3d} to regularize the output geometry. The overall objective is defined as the weighted sum of these losses.
We use an Adam optimizer~\citep{kingma2014adam} with a learning rate of $0.01$ to update all parameters for 1500 iterations in stage 1 and 500 iterations in stage 2, taking roughly 20-25 minutes per object.


\section{Ablative Analyses}
\label{sec:ablation}

\subsection{Novel-view video synthesis}
In~\cref{tab:nvs-ablation}, we report the ablative results of SV4D 2.0 model on the ObjaverseDy dataset~\cite{deitke2023objaverse,xie2024sv4d}, which justify our design choices and contributions.

\inlinesection{Optional reference multi-views.}
We first show that a fully multi-view unconditional model (w/o reference multi-views) performs considerably worse since it cannot leverage previous generation as multi-view condition to maintain temporal coherence between multiple generations.
Next, we show that using VAE latents of reference multi-views produces better results than using CLIP embedding, as VAE latents include more detailed spatial features whereas CLIP embedding contains high-level semantic and global texture information.

\inlinesection{3D attention and camera embedding conditioning.}
To ablate the network architecture, we show that using view attention in SV4D~\cite{xie2024sv4d} (instead of 3D attention) and adding view indices as conditioning (instead of camera embedding) leads to worse image quality and multi-view consistency, especially when generating sparse novel views.

\inlinesection{Improved data curation.}
We also report the model performance without using the improved data in ObjaverseDy++ for training, which further strengthens the effectiveness of our data curation approaches.

\inlinesection{Progressive 3D-4D training.}
Finally, we show the importance of progressive 3D-to-4D training by evaluating the model directly trained on 4D data.

\subsection{4D optimization}
In~\cref{tab:4d-ablation}, we show the ablative results of our 4D optimization method on the ObjaverseDy dataset~\cite{deitke2023objaverse,xie2024sv4d}. In particular, we show that the proposed stage-2 refinement significantly improves all image and video metrics. The orthogonal view and progressive frame sampling also leads to better image quality (LPIPS, CLIP-S, PSNR) and temporal consistency (FVD-F) compared to random view and frame sampling.


\begin{figure*}[t!]
    \centering
    \includegraphics[width=.65\linewidth]{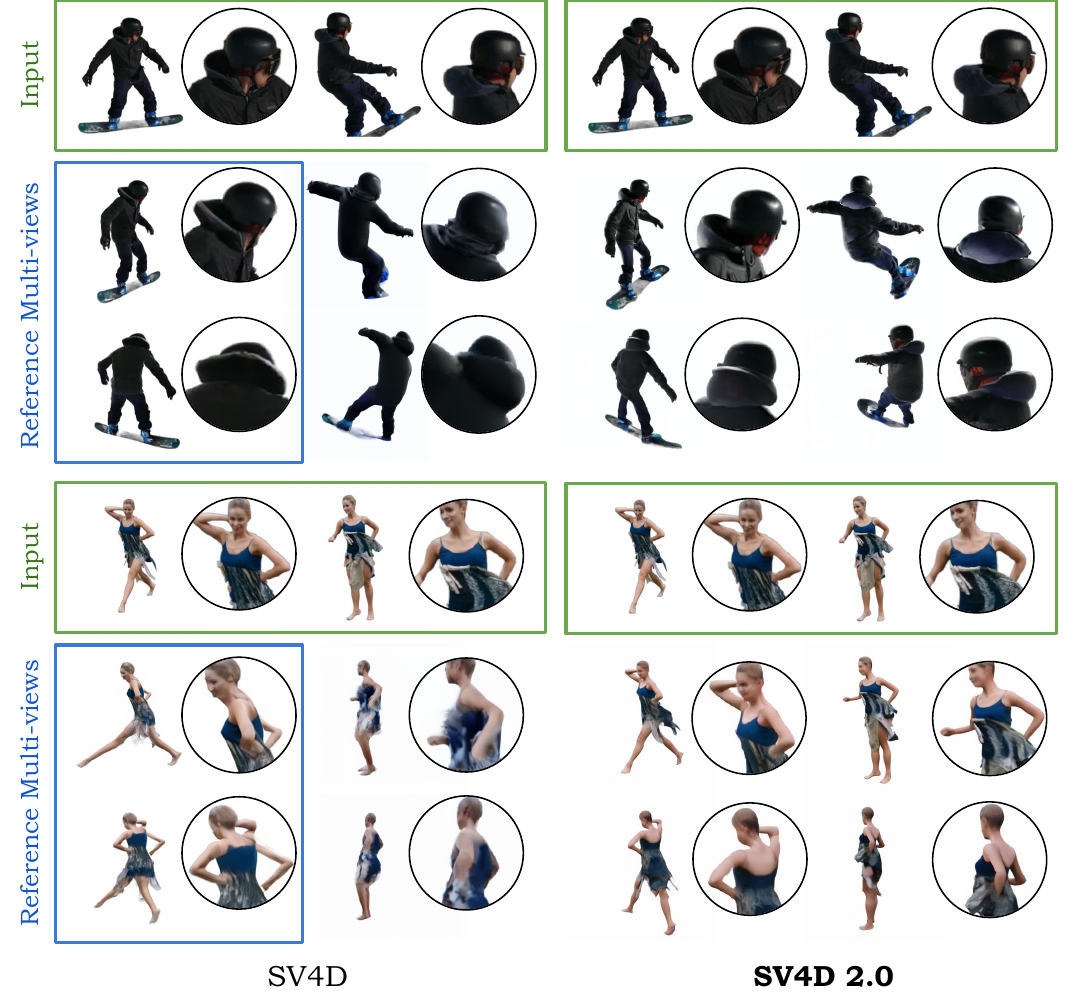}
    \vspace{-2mm}
    \caption{
        \textbf{Dependency of reference multi-views.} \newadd{SV4D~\cite{xie2024sv4d} relies on the reference multi-views produced by SV3D~\cite{voleti2024sv3d}, which often conflicts with the later frames of the input video (\eg, jacket hood of the snowboarder and three arms of the dancing women) and leads to blurry outputs. In contrast, SV4D 2.0 can better leverage the information in all input frames to produce sharper and more faithful details.}
    }
\label{fig:conflicting_ref}
\end{figure*}

\subsection{Limitations}
Since we follow SV3D~\cite{voleti2024sv3d} and SV4D~\cite{xie2024sv4d} to use elevation and azimuth (2 degree of freedom) as our camera parametrization, the model cannot handle videos with strong perspective distortion.
Additionally, SV4D 2.0 focuses on object-centric videos, limiting its applicability to general video applications like dynamic 3D scene generation. Generalizing the model to scene-level videos poses an interesting future work.

\section{Additional Results}
\label{sec:additional_results}

\subsection{Conflicting reference multi-views}
\newadd{To demonstrate the importance of removing dependency on reference multi-views, we show some examples of reference multi-views conflicting with the input videos in~\cref{fig:conflicting_ref}.  Such cases often occur when the subject occludes itself (right) or shows different views in the video due to rotation or camera motion (left). This forces the model to merge conflicting information in the input video and reference multi-views, resulting in blurry details as shown in SV4D outputs.}

\subsection{Robustness to occlusion}
\begin{figure*}[t!]
    \centering
    \includegraphics[width=.65\linewidth]{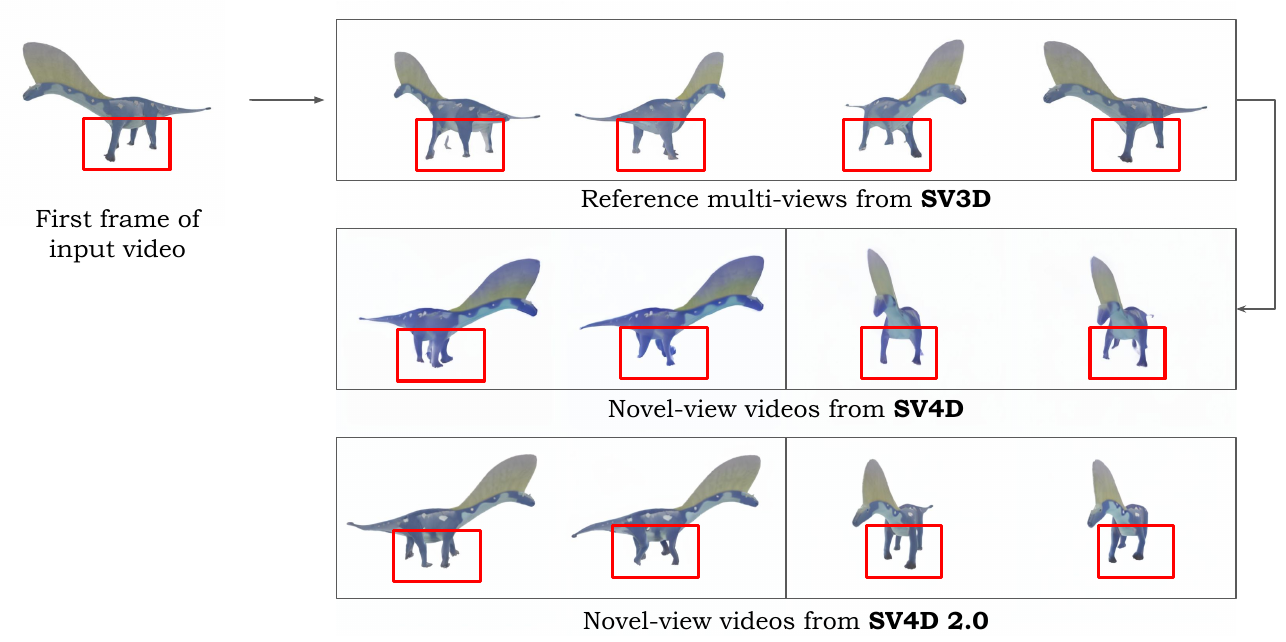}
    \vspace{-2mm}
    \caption{
        \textbf{Robustness to occlusion.} SV4D~\cite{xie2024sv4d} relies on the reference multi-views generated by SV3D~\cite{voleti2024sv3d}, conditioning on the first frame of the input video. We show that it suffers from self-occlusion in the first frame and often produces results with blurry artifacts or missing parts. In contrast, SV4D 2.0 is more robust to such occlusion by jointly taking multiple input frames to generate the novel-view videos.
    }
\label{fig:occlusion}
\end{figure*}

In~\cref{fig:occlusion}, we further show an example of self-occlusion in the input video. Since the reference multi-view generation by SV3D~\cite{voleti2024sv3d} is only conditioned on the first frame of the input video, it fails to capture all 4 legs in several novel views. This results in inconsistent or blurry novel-view videos synthesized by SV4D~\cite{xie2024sv4d}. In contrast, SV4D 2.0 generates consistent details by taking in multiple input frames as condition, showing higher robustness to occlusion.

\begin{figure*}[t!]
    \centering
    \includegraphics[width=.95\linewidth]{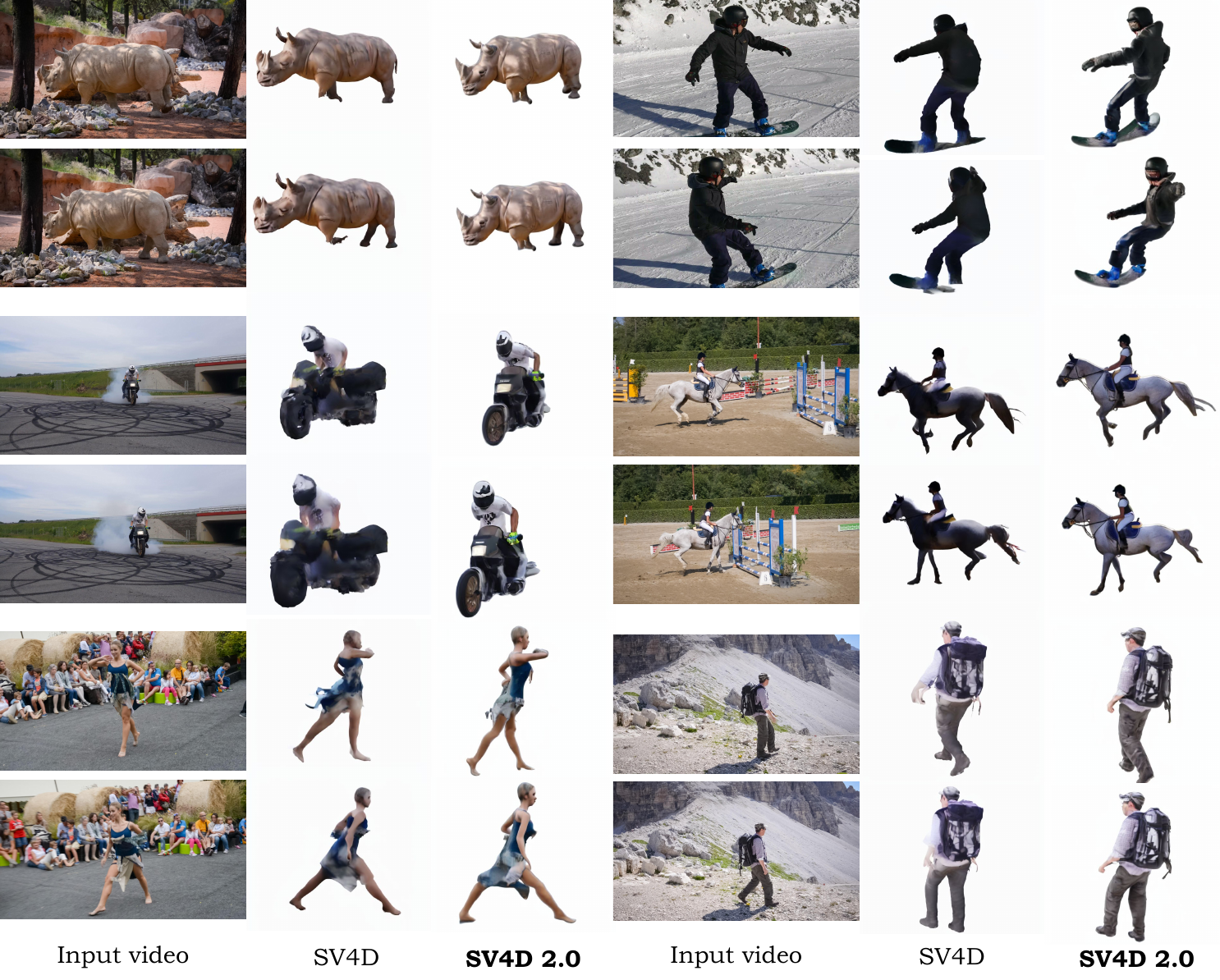}
    \vspace{-2mm}
    \caption{
        \textbf{Additional results of novel-view video synthesis on DAVIS~\cite{Caelles_arXiv_2019}.} We show that SV4D 2.0 can generalize better to the real-world data and produce higher-fidelity videos compared to SV4D~\cite{xie2024sv4d}.
    }
\label{fig:results_nvs_2}
\end{figure*}

\begin{figure*}[t!]
    \centering
    \includegraphics[width=\linewidth]{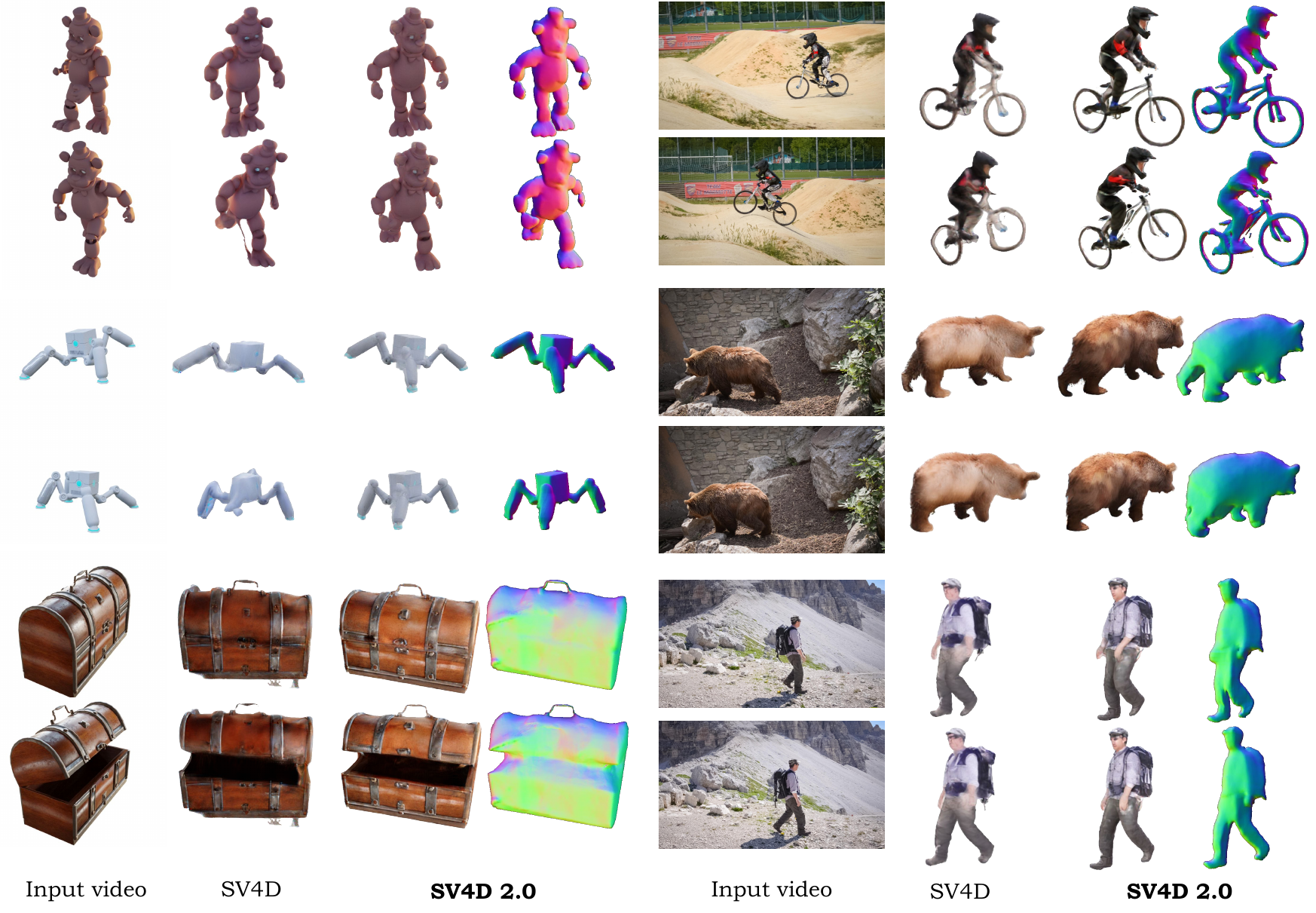}
    \caption{
        \textbf{Additional results of 4D optimization on ObjaverseDy~\cite{deitke2023objaverse,xie2024sv4d} (left) and DAVIS~\cite{Caelles_arXiv_2019} (right).} We show the RGB renders of 4D outputs by SV4D~\cite{xie2024sv4d} and SV4D 2.0, as well as our rendered normals. By comparison, our 4D assets can better capture the intricate geometry and texture details during motion.
    }
\label{fig:results_4d_2}
\end{figure*}

\subsection{Novel-view video synthesis}

We show additional NVVS results on the real-world DAVIS~\cite{Caelles_arXiv_2019} videos in~\cref{fig:results_nvs_2}, demonstrating higher-fidelity details and better generalization capability of SV4D 2.0 compared to SV4D~\cite{xie2024sv4d}.

\subsection{4D optimization}

In~\cref{fig:results_4d_2}, we show additional 4D results on the ObjaverseDy~\cite{deitke2023objaverse,xie2024sv4d} and DAVIS~\cite{Caelles_arXiv_2019} videos. The RGB and normal renders demonstrate that SV4D 2.0 can capture higher-quality texture and geometry details in large motion compared to SV4D~\cite{xie2024sv4d}.


\subsection{4D representation: DyNeRF \textit{vs} 4D Gaussian}
\begin{figure*}[t!]
    \centering
    \includegraphics[width=.97\linewidth]{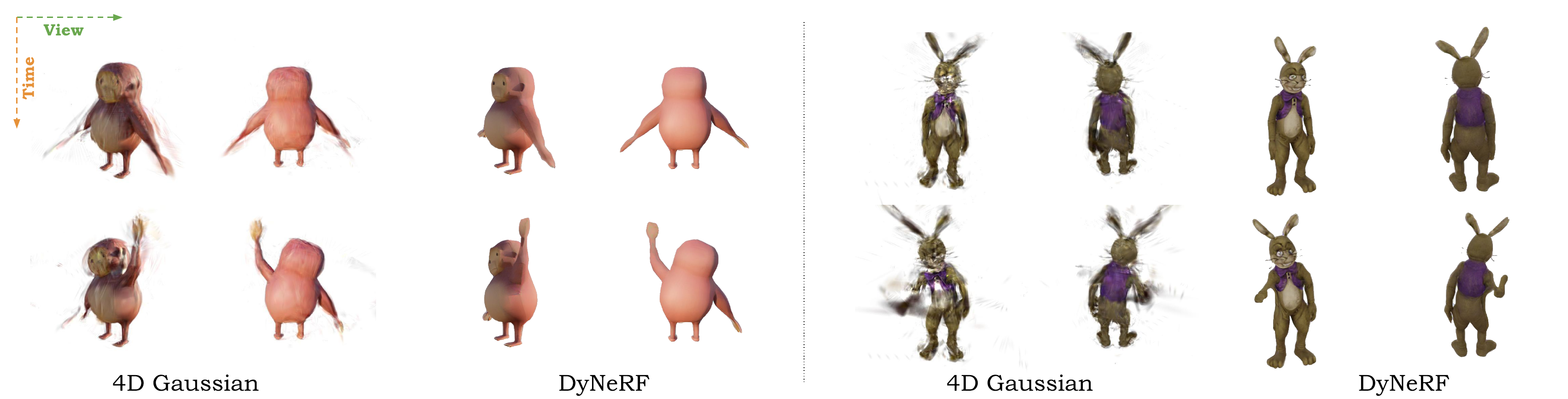}
    \vspace{-2mm}
    \caption{
        \newadd{
        \textbf{4D Optimization with 4D Gaussians.}
        4D Gaussians suffer from temporal flickering and blurry artifacts due to its discrete nature, while Dy-NeRF interpolates better across sparse views and fast motion.
        }
    }
\label{fig:4d_gaussian}
\end{figure*}

\newadd{
We show some results using 4D Guassians in \cref{fig:4d_gaussian}. 
As mentioned in SV4D~\cite{xie2024sv4d}, 4D Gaussians suffer from temporal flickering and blurry artifacts due to its discrete nature, while Dy-NeRF interpolates better across sparse views and fast motion.
}

\subsection{More L4GM results}
\begin{figure*}[t!]
    \centering
    \includegraphics[width=.98\linewidth]{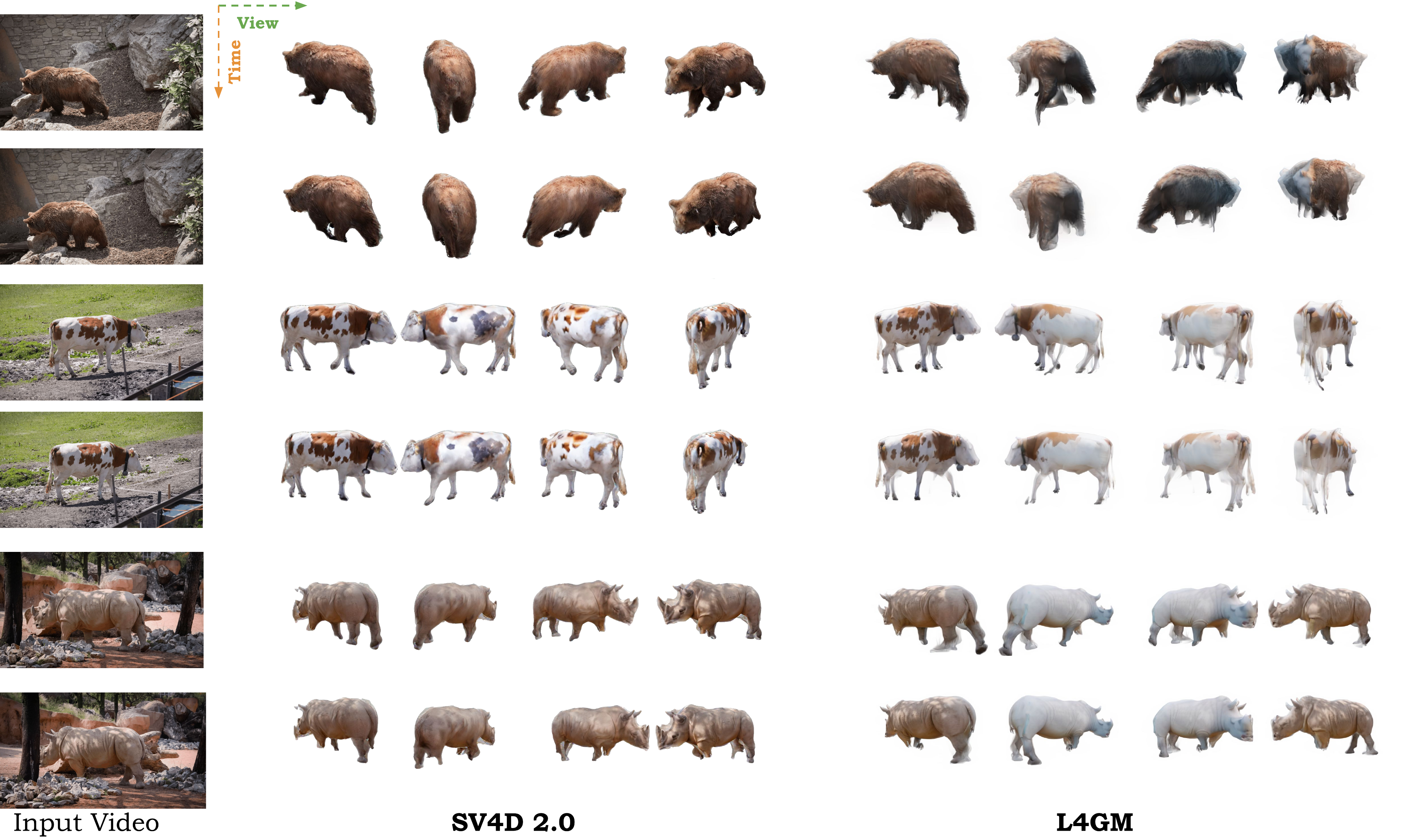}
    \vspace{-2mm}
    \caption{
        \newadd{\textbf{Comparing to L4GM with real-world video input.}
        Compared to the L4GM, \name generalizes better with real-world inputs because we leverage video priors by resuming pre-trained weights from the video generative model.}
    }
\label{fig:l4gm_real}
\end{figure*}

\begin{figure*}[t!]
    \centering
    \includegraphics[width=.97\linewidth]{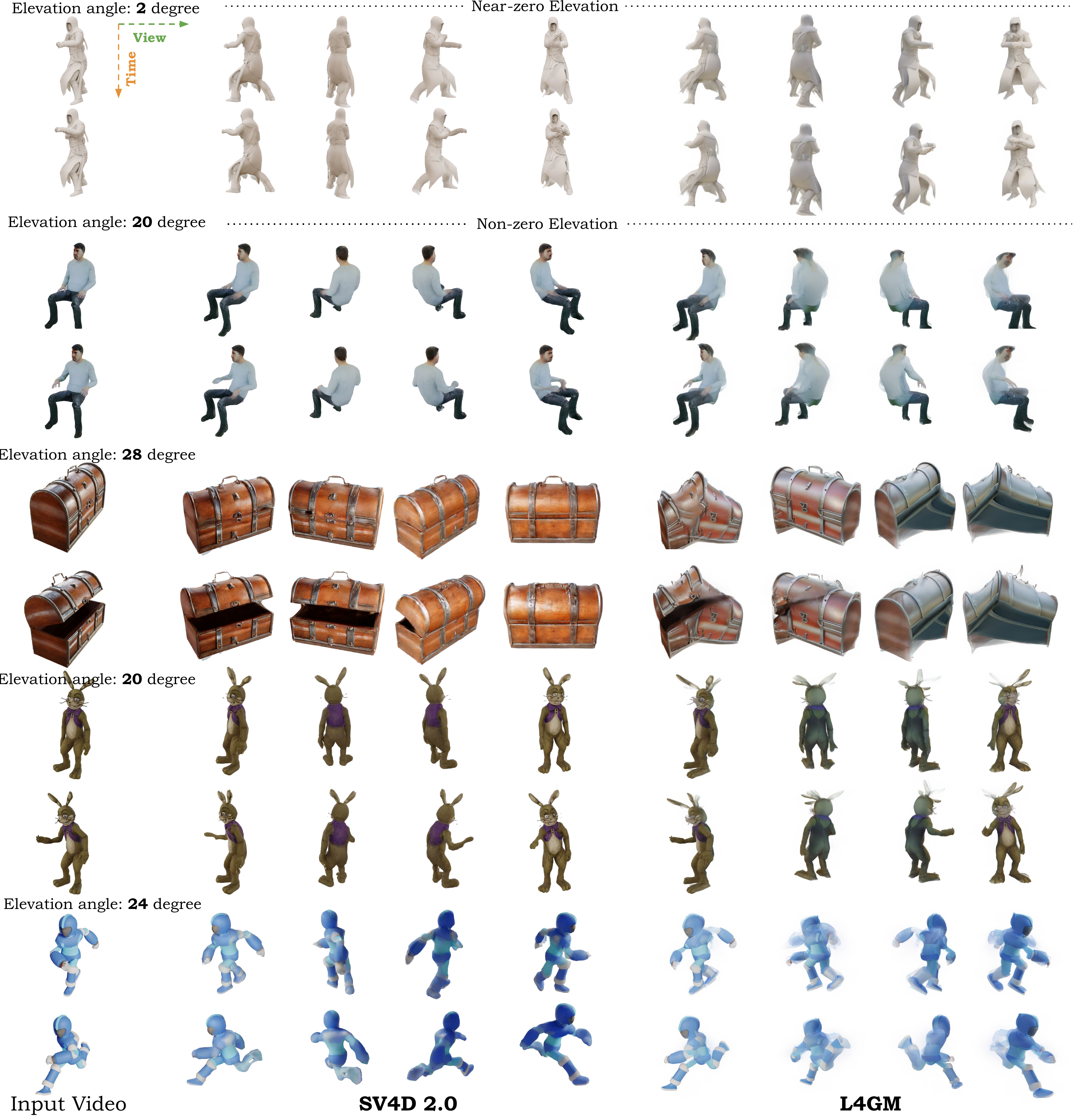}
    \caption{
        \newadd{\textbf{Comparing to L4GM with non-zero elevation video input.} Compared to the L4GM, \name generalizes much better on non-zero elevation inputs. L4GM was trained with only 0\degree elevation videos.
        }
    }
\label{fig:l4gm_elev}
\end{figure*}

\newadd{
We show more comparisons with L4GM on real-world videos in \cref{fig:l4gm_real} and video with non-zero elevation in \cref{fig:l4gm_elev}. 
L4GM does not generalize well on real-world data (no video prior like ours) and struggles with videos at non-zero elevation (training data primarily at 0° elevation).
}


\subsection{Efficiency analysis.}
\newadd{
Our model takes roughly 1 minute to generate 
4 videos of 12 frames at 512 $\times$ 512 resolution on a H100 GPU, and 21 minutes for 4D optimization. 
It requires up to 30GB of memory.
}

\section{Supplementary Video}
In addition to the paper and appendix, we also include a comprehensive video in the supplementary material, providing an in-depth introduction to our framework and more visual results to demonstrate the effectiveness of SV4D 2.0.

\end{document}